\documentclass[final]{cvpr}

\usepackage{times}
\usepackage{epsfig}
\usepackage{graphicx}
\usepackage{amsmath}
\usepackage{amssymb}
\usepackage{comment}
\usepackage{multirow}
\usepackage{subfigure}
\usepackage[symbol]{footmisc}

\usepackage[ruled,linesnumbered]{algorithm2e}

\usepackage[pagebackref=true,breaklinks=true,colorlinks,bookmarks=false]{hyperref}
\newcommand{\tabincell}[2]{\begin{tabular}{@{}#1@{}}#2\end{tabular}}  
\newcommand{\red}[1]{{\color{red}{#1}}}
\newcommand{\blue}[1]{{\color{blue}{#1}}}

\def\cvprPaperID{10845} 
\def\confYear{CVPR 2021}

\begin{document}

\title{Prioritized Architecture Sampling with Monto-Carlo Tree Search}

\author{Xiu Su$^{1}$\thanks{Equal contributions.}, Tao Huang$^{2*}$, Yanxi Li$^{1}$, Shan You$^{2,3}$\thanks{Corresponding authors.}, \\
Fei Wang$^{2}$, Chen Qian$^{2}$, Changshui Zhang$^{3}$, Chang Xu$^{1\dagger}$\\
$^1$School of Computer Science, Faculty of Engineering, The University of Sydney, Australia\\
$^2$SenseTime Research\\
$^3$Department of Automation, Tsinghua University,\\
Institute for Artificial Intelligence, Tsinghua University (THUAI), \\
Beijing National Research Center for Information Science and Technology (BNRist) \\
{\tt\small xisu5992@uni.sydney.edu.au, \{huangtao,youshan,wangfei,qianchen\}@sensetime.com}\\
{\tt\small yali0722@uni.sydney.edu.au, zcs@mail.tsinghua.edu.cn, c.xu@sydney.edu.au}
}

\maketitle

\begin{abstract}
    One-shot neural architecture search (NAS) methods significantly reduce the search cost by considering the whole search space as one network, which only needs to be trained once. However, current methods select each operation independently without considering previous layers. Besides, the historical information obtained with huge computation cost is usually used only once and then discarded. In this paper, we introduce a sampling strategy based on Monte Carlo tree search (MCTS) with the search space modeled as a Monte Carlo tree (MCT), which captures the dependency among layers. Furthermore, intermediate results are stored in the MCT for future decisions and a better exploration-exploitation balance. Concretely, MCT is updated using the training loss as a reward to the architecture performance; for accurately evaluating the numerous nodes, we propose node communication and hierarchical node selection methods in the training and search stages, respectively, which make better uses of the operation rewards and hierarchical information. Moreover, for a fair comparison of different NAS methods, we construct an open-source NAS benchmark of a macro search space evaluated on CIFAR-10, namely NAS-Bench-Macro. Extensive experiments on NAS-Bench-Macro and ImageNet demonstrate that our method significantly improves search efficiency and performance.
    For example, by only searching $20$ architectures, our obtained architecture achieves $78.0\%$ top-1 accuracy with 442M FLOPs on ImageNet. Code (Benchmark) is available at: \url{https://github.com/xiusu/NAS-Bench-Macro}.
 
\end{abstract}

\begin{figure}[tbp]
    \begin{center}
        \includegraphics[width=0.98\linewidth]{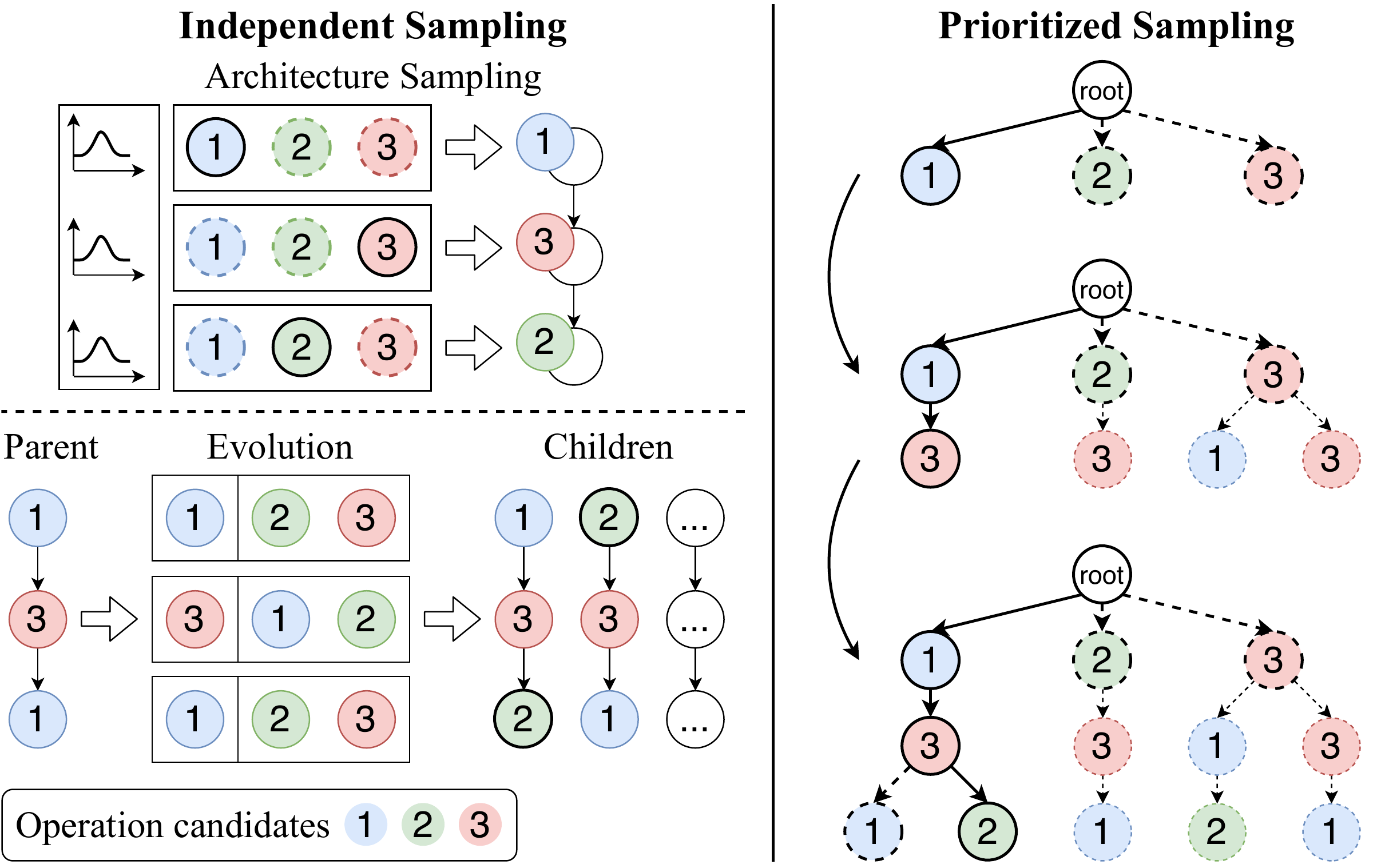}
    \end{center}
    \caption{Comparison between existing methods (left) and our method (right). The existing method treats each layer independently in training (top-left) and search (bottom-left) stages, while our method models the search space with dependencies to a unified tree structure.}
    \label{fig:motivation}
    \vspace{-4mm}
\end{figure}

\section{Introduction}

Deep learning has not only thrived in various tasks as image recognition and object detection \cite{Liao_2020_CVPR,wei2020point,han2021transformer}, but also achieved remarkable performance on mobile edge devices \cite{mobilenetv2,hinton2015distilling,you2017learning,tang2020reborn,du2020agree,han2020ghostnet}. Neural architecture search (NAS) makes a step further; it even liberates the reliance on expert knowledge and obtains higher performance by developing more promising and cost-effective architectures \cite{tang2020semi,su2021locally,guo2020hit}. Despite the inspiring success of NAS, the search space of conventional NAS algorithms is extremely large, leading the exhaustive search for the optimal network will be computation prohibited. To accommodate the searching budget, heuristic searching methods are usually leveraged and can be mainly categorized into reinforcement learning-based \cite{mnasnet,tan2019efficientnet}, evolution-based \cite{guo2020single,you2020greedynas}, Bayesian optimization-based \cite{zhou2019bayesnas,wang2020learning}, and gradient-based methods \cite{liu2018darts,proxylessnas,yang2021towards,huang2020explicitly}.

\begin{figure*}[tbp]
    \begin{center}
        \includegraphics[width=0.92\linewidth]{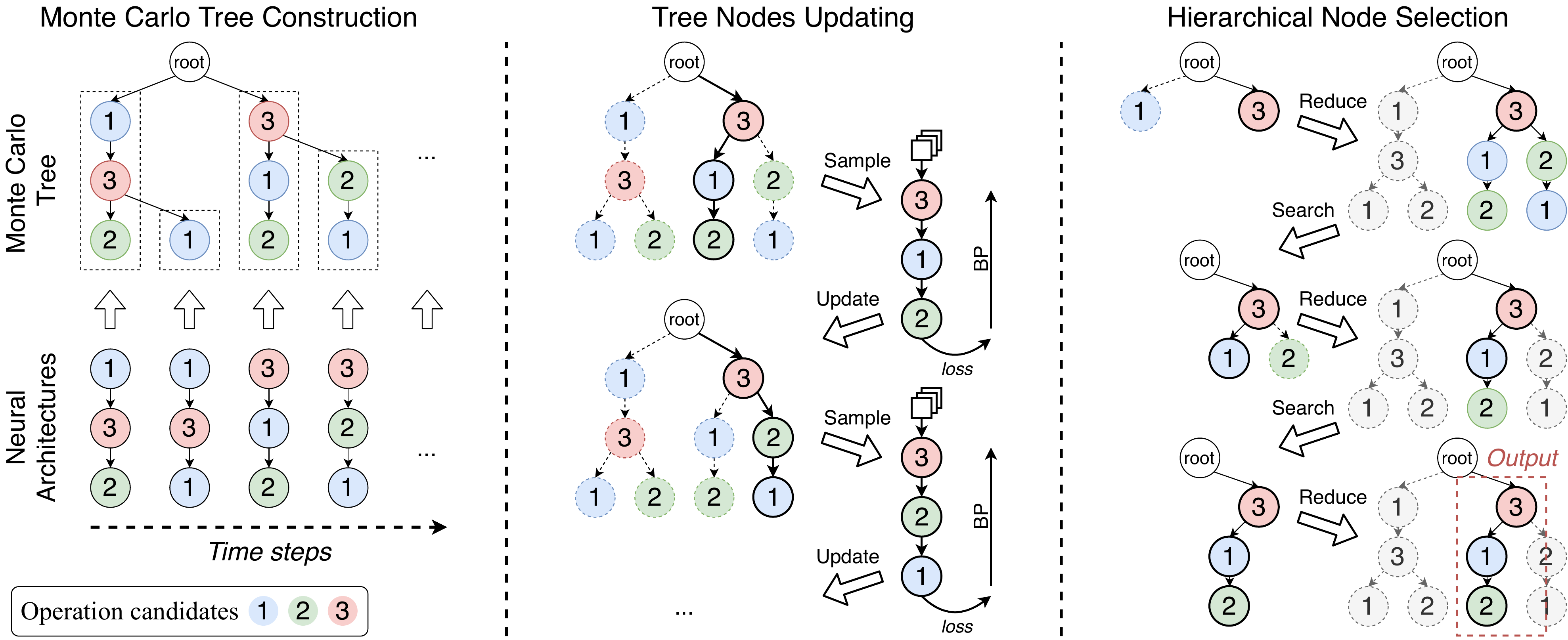}
    \end{center}
    \caption{The overall framework of MCT-NAS, which models the search space into a MCT (left), then updates the tree with a prioritized sampling strategy during training (middle), finally searches the optimal architecture using hierarchical node selection (right). 
    }
    \label{fig:framework}
    \vspace{-4mm}
\end{figure*}

Among these methods, one-shot NAS methods enjoy significant efficiency since they only train the whole search space(\ie, supernet) simultaneously. In current mainstream methods of one-shot NAS, a supernet is usually considered as a performance evaluator for all architectures within it, and the optimal architecture is obtained by evaluating the architectures with a validation set. Since it is unrealistic to evaluate all the architectures in the search space, current methods usually deal with a restricted search number, \eg, $1000$ vs. $13^{21}$.
For the sake of efficiently searching a good architecture with this limited search number, several heuristic search methods have been developed \cite{guo2020single, you2020greedynas}, \eg, evolutionary algorithms (EA), Bayesian optimization. 

Though existing one-shot NAS methods have achieved impressive performance, they often consider each layer separately while ignoring the dependencies between the operation choices on different layers, which leads to an inaccurate description and evaluation of the neural architectures during the search. For example, Gaussian Processes (GP) in Bayesian optimization requires that the input attributes (OPs) are independent of each other 
\cite{zhou2019bayesnas,wang2020learning}, and the cross mutations of OPs in evolutionary search are often carried out separately in each layer \cite{guo2020single,you2020greedynas}. In fact, for a feed-forward neural network, the choice of a specific layer relates to its previous layers and contributes to its post layers. 

In this paper, we highlight the dependencies between operations on different layers through establishing a Monte Carlo Tree (MCT) in the architecture search space and develop an effective sampling strategy based on Monte Carlo Tree Search (MCTS) for NAS (see Figure \ref{fig:motivation}),
The training loss is considered as a reward representation of each node (OP) in MCT, which is used for determining which architecture to be explored. Meanwhile, for a better evaluation of numerous posterior nodes, we propose a node communication technique to share the rewards among nodes with the same operation and depth. The dependencies between different operations on different layers can be accurately modeled with MCT. During searching on the supernet, to evaluate the nodes more accurately, we propose a hierarchical node selection of MCT, which hierarchically updates the rewards on those less-visited nodes using validation data.

For a better comparison between different NAS methods, we propose a NAS benchmark on macro search space named \textit{NAS-Bench-Macro}, which trains $6561$ networks isolatedly on CIFAR-10 dataset. Experiments on NAS-Bench-Macro show our superiority in efficiently searching the optimal architectures. We also implement our MCTS-NAS on the large-scale benchmark dataset ImageNet \cite{russakovsky2015imagenet} with extensive experiments. Under the same FLOPs budgets or acceleration, our method significantly improves the search efficiency with better performances compared to other one-shot NAS methods \cite{guo2020single, you2020greedynas}.
For example, we decrease the search number of sub-networks (subnets) from 1000 to 20, which reduces a large amount of search cost.
Besides, the obtained architecture achieves 78.0\% Top-1 accuracy with the MobileNetV2 search space with only 442M FLOPs.  

\section{Related work}
\textbf{One-shot NAS.} 
The recent emerging one-shot approaches significantly reduce the search cost of NAS by considering architectures as sub-graphs of a densely connected supernet.
DARTS \cite{liu2018darts,yang2020ista} relaxes the discrete operation selection as a continuous probability distribution, which can be directly optimized with gradient descent.
Training all operation candidates in parallel is memory-consuming and computationally inefficient.
To address such a challenge, Single Path One-Shot \cite{guo2020single} proposes to train each architecture alternately and adopt a two-stage search schema, where network training is first performed by uniform path sampling. Then the final architecture is searched using a validation dataset.

Targeting at minimizing the evaluation gap between the weight-sharing subnet and standalone network,
GreedyNAS \cite{you2020greedynas} introduces a progressive search space reduction strategy.  A greedy path filtering technique is introduced to let the supernet pay more attention to those potentially-good paths. Similarly, SGAS \cite{li2020sgas} propose to prune operation candidates in the search space in a greedy manner. With the low-ranked candidates being removed, the search space is reduced, and the supernet can focus on the remaining ones for sufficient training and proper evaluation.

The greedy search \cite{you2020greedynas} can be arbitrary and might be trapped by the local optimum.
In contrast, the evolutionary algorithm \cite{deb2002fast} works as a heuristic search method that leverages the prior information of the searched subnets that have achieved profound success in NAS. For example, Single Path One-shot \cite{guo2020single} searched the optimal subnets from supernet with the evolutionary algorithm. Afterward, many algorithms \cite{you2020greedynas,chu2019scarletnas} follow this strategy by leveraging the evolutionary algorithm to search for optimal subnets from the designed supernet. However, the evolutionary algorithm only allows mutation and crossover over operations, which fails to consider the relations over layers, leading to sub-optimal results.

To solve this issue, many methods involve tree-based MCTS \cite{negrinho2017deeparchitect, wang2019sample, wang2020learning, wang2020neural} into NAS. However, these methods only explore MCTS as a simple sampler with a huge search space. For example, the AlphaX \cite{wang2020neural} trains each searched architecture independently for evaluation, which is computationally expensive. LaNAS \cite{wang2019sample} only adopts MCTS on the search stage, while on training, it trains the supernet with random masks. Compared to the above methods, with a macro one-shot search space, our MCT-NAS stores much more information for efficient and accurate searching on both the training and search stage, promoting an accurate evaluation of the numerous nodes.

\section{Methodology}

In this paper, we conduct an architecture search on a macro NAS search space, in which the searching layers are stacked sequentially, and each layer selects one operation from the operation set $\mathcal{O} = \{o_i\}$ with size $N$.
With a search space $\mathcal{A}$ of $L$ layers, an architecture $\boldsymbol{\alpha} \in \mathcal{A}$ can be uniquely represented by a set of operations, where $\boldsymbol{\alpha} = \{o^{(l)}\}_{l \in \{1, \dots, L\}}$.

To search for the optimal architecture $\boldsymbol{\alpha}^* \in \mathcal{A}$, as illustrated in Figure \ref{fig:framework}, we use a two-stage procedure consists of training and search. First, in the training stage (see the left and middle sub-figures in Figure \ref{fig:framework}), we sample and train architectures in the search space alternately. Different from the uniform sampling strategy adopted in previous works, we sample architectures with the help of the \textit{Monte Carlo tree} (MCT), which well balances the exploration and exploitation of the search space, and the training loss of each architecture is stored in the MCT for future decision. Second, after the supernet is well trained, we adopt a node communication strategy to evaluate the less-visited nodes in the constructed MCT using a validation set; thus, the nodes can be searched more efficiently and accurately in the search stage. Then we search architectures using a hierarchical node selection method and then obtain the final architecture with the highest validation accuracy.

\subsection{Search Space Modeling with MCT}
In one-shot NAS, the over-parameterized weight-sharing supernet is usually trained with a sampling strategy, in which only one subnet will be sampled and optimized each iteration. While at the search stage, the subnets are also sampled and evaluated standalone. So the sampling strategy highly determines the performance of obtained architecture. Existing NAS methods treat different layers independently on sampling; however, in this paper, we highlight the dependency modeling and propose to sample the subnets from a MCT-based distribution.

We first analyze the common sampling strategy in existing NAS methods that select each operation independently. Therefore, 
the probability distribution of sampling an architecture $\boldsymbol{\alpha}$ can be formulated as
\begin{equation} \label{eq:prioritized-sampling:independent}
    P(\boldsymbol{\alpha}) = P(o^{(1)}, \dots, o^{(L)}) = \prod_{l=1}^{L} P(o^{(l)}),
\end{equation}
where $P(o^{(l)})$ denotes the probability distribution of the operation selection in the layer $l$.
In Eq. \eqref{eq:prioritized-sampling:independent}, the probability of selecting an operation is solely determined by the layer $l$ independently.
However, we argue that in a chain-structured network, the selection of operation at each layer should depend on operations in the previous layers.

To capture the dependency among layers and leverage the limited combinations of operations for better understanding of the search space, we replace $P(o^{(l)})$ in Eq.\eqref{eq:prioritized-sampling:independent} with a conditional distribution for each $2 \leq l \leq L$.
Therefore, we reformulate Eq. \eqref{eq:prioritized-sampling:independent} as follows:
\begin{equation}
\begin{aligned}
    P(\boldsymbol{\alpha}) &= P(o^{(1)}, \dots, o^{(L)})\\
    &= P(o^{(1)}) \cdot \prod_{l=2}^{L} P(o^{(l)} | o^{(l)}, \dots, o^{(l-1)}),
    \label{eq:prioritized-sampling:dependent}
\end{aligned}
\end{equation}
where $P(o^{(l)} | o^{(1)}, \dots, o^{(l-1)})$ is the conditional probability distribution of the operation selection in the layer $l$ conditioned on its previous layers $1$ to $l-1$. Note that $l=1$ has no previous layer, so $P(o^{(1)})$ is still independent.

Inspired by Eq.\eqref{eq:prioritized-sampling:dependent}, we find this conditional probability distribution of search space can be naturally modeled into a tree-based structure, and the MCTS is targeting this structure for a better exploration-exploitation trade-off. As a result, we propose to model the search space with a MCT $\mathcal{T}$.
In MCT, each node $v_i^{(l)} \in \mathcal{T}$ corresponds to selecting an operation $o_i^{(l)} \in \mathcal{O}$ for the layer $l$ under the condition of its ancestor nodes, so the architecture representation $\boldsymbol{\alpha} = \{o^{(l)}\}_{l \in \{1, \dots, L\}}$ can also be uniquely identified in the MCT.
As Figure \ref{fig:framework} shows, the architectures are independently represented by paths in the MCT, and different choices of operations lead to different child trees; thus, the dependencies of all the operation selections can be naturally formed.

\subsection{Training with Prioritized Sampling} 

With the MCT $\mathcal{T}$, we can perform a prioritized sampling of architectures and store the searching knowledge in it. For each node $v_i^{(l)}$, there are two values stored in $\mathcal{T}$, including a Q-value $Q(v_i^{(l)})$ measuring the quality of selection and a number of visits $n_i^{(l)}$ counting how many times this selection occurs. For the sake of efficiency, we use the training loss $\mathcal{L}_{tr}$ as a representation of selection quality (Q-value).
However, since the supernet's network weights are continuously optimized during the search, the performance of a certain architecture will be enhanced along with the optimization procedure.
As a result, it is unfair to directly compare architectures evaluated at different iterations. An intuitive and effective way to solve this issue is to introduce a normalized performance ranking factor.
We use a moving average of losses $\widetilde{\mathcal{L}}_t$ as the baseline:
\begin{equation}
    \widetilde{\mathcal{L}}_t = \beta \cdot \widetilde{\mathcal{L}}_{t-1} + (1 - \beta) \cdot \mathcal{L}_{tr}(\boldsymbol{\alpha}_t),
    \label{eq:tree-update:moving-average}
\end{equation}
where $\beta \in [0, 1]$ is the reduction ratio of the moving average, and $\mathcal{L}_{tr}(\boldsymbol{\alpha}_t)$ denotes the training loss of the current architecture $\boldsymbol{\alpha}_t$ in iteration $t$.  $\widetilde{\mathcal{L}}_t$ represents the convergence status of supernet at iteration $t$, so we can compare $\mathcal{L}_{tr}(\boldsymbol{\alpha}_t)$ with it to get a relative performance of the architecture, thus the Q-value of the selected nodes are updated with,
\begin{equation}
    Q (v_i^{(l)}) = \frac{\widetilde{\mathcal{L}}_t}{\mathcal{L}_{tr}(\boldsymbol{\alpha}_t)}.
    \label{eq:tree-update:q-value}
\end{equation}

In the sampling, for a better exploration-exploitation balance, we select node in the MCT based on the \textit{Upper Confidence Bounds for Trees} (UCT) \cite{kocsis2006bandit}.
Given a parent node $v_p^{(l-1)} \in \mathcal{T}$ with a number of visits $n_p^{(l-1)}$, we select its child nodes according to the UCT function.
The UCT function of a child node $v_i^{(l)} \in \mathcal{T}$ is calculated by
\begin{equation}
    \operatorname{UCT}(v_i^{(l)}) = \frac{Q(v_i^{(l)})}{n_i^{(l)}} + C_1 \sqrt{\frac{\log (n_p^{(l-1)})}{n_i^{(l)}}},
    \label{eq:uct-func}
\end{equation}
where $C_1 \in \mathbb{R}_+$ is a constant controlling the trade-off between exploration and exploitation.

In general, only the node with the highest UCT score will be selected by MCT, but it prevents the sampling method from exploring more diverse architectures.  
Instead of directly selecting the node with the maximum UCT function, we propose MCT-NAS to relax the operation selection in one layer to a probability distribution using softmax function, \ie,
\begin{equation}
    P_t(v_i^{(l)}) = \frac{\exp \left(\operatorname{UCT}(v_i^{(l)}) / \tau \right)}{\sum_{j\le N^l} \exp \left(\operatorname{UCT}(v_j^{(l)}) / \tau \right)},
    \label{eq:softmax-sampling}
\end{equation}
where $N^l$ denotes the total node number in depth $l$, and $\tau$ is a temperature term. Note that when $\tau \to 0$, it becomes an approximated categorical distribution that almost always selects the operation with the maximal UCT score. We set $\tau$ to $0.0025$ in all of our experiments.

In the end, during training, the sampling distribution is changed from uniform distribution to our prioritized sampling distribution. We investigate the dependence between operations on different layers and promote the exploration and exploitation of good architectures. Note that, in this paper, we also use uniform sampling at the beginning of the training for a warm-up start.

\subsection{Node Communication}

\begin{figure}[tbp]
    \begin{center}
        \includegraphics[width=1.0\linewidth]{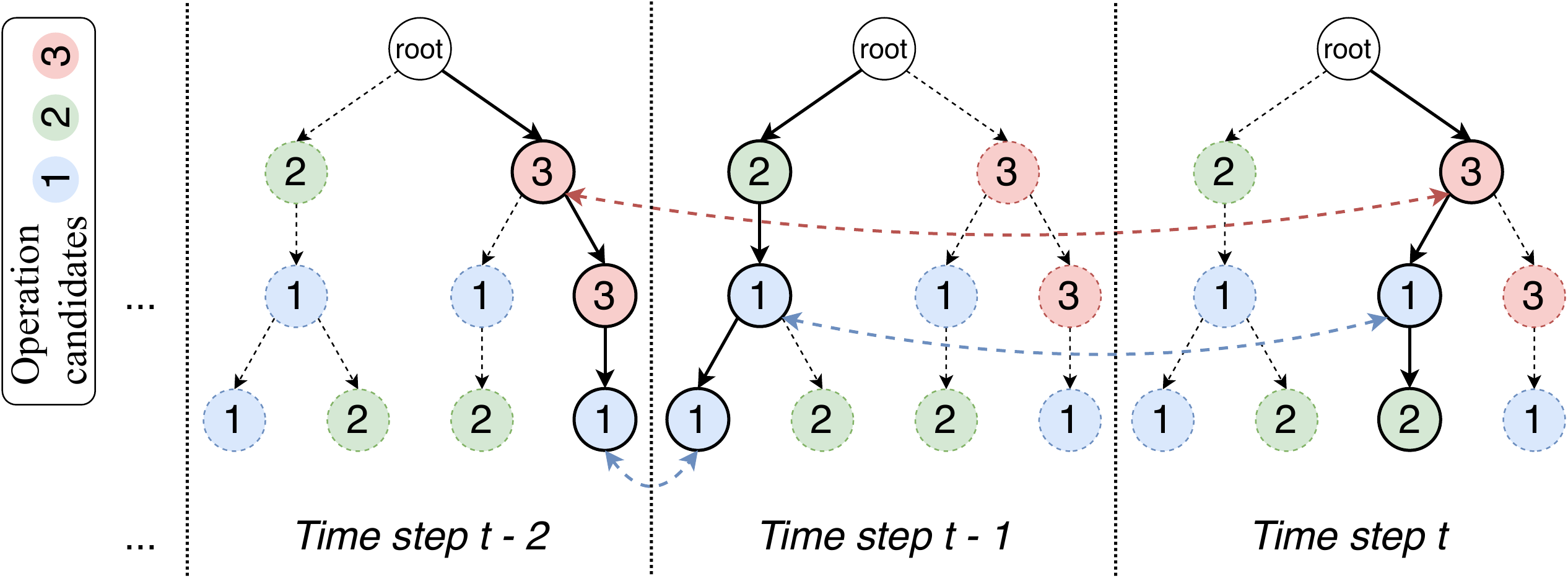}
    \end{center}
    \vspace{-4mm}
    \caption{Node communication with the same selected operation.}
    \vspace{-4mm}
    \label{node communication}
\end{figure}

In MCT, each architecture in the search space corresponds to a unique path. As the increment of the depth, the number of nodes grows exponentially. For example, the MCT will have $N^L$ leaf nodes with a search space of size $N^L$. It is, therefore, impossible to explore all these numerous nodes. However, as in the supernet, the same operations in a layer share the same weights, which inspires that the nodes should have some common knowledge from their operation type. For a better reward representation on the nodes in posterior depth, we propose a \textit{node communication} technique to share the rewards for nodes with the same operation and depth.

Concretely, to represent the reward of an operation in a specific layer, we use a moving average of all the rewards of its corresponding nodes, denoted as node communication score $G$ in Figure \ref{node communication}.
For each operation in layer $l$, denoted as $o_j^{(l)}$, its score $G$ is updated by the Q-value of corresponding nodes $v_i^{(l)}$ , \ie,  
\begin{equation}
    G(o_j)^{(l)} \gets \gamma \cdot G(o_j)^{(l)} + (1 - \gamma) \cdot Q (v_i^{(l)}),
    \label{eq:node-comm:generalized-quality}
\end{equation}
where $\gamma \in [0, 1]$ is a reduction ratio.

As a result, the reward of a node can be jointly represented by the Q-value of this node, and the node communication scores $G$ of its corresponding operation. By adding Eq.\eqref{eq:node-comm:generalized-quality}, the UCT function in Eq.\eqref{eq:uct-func} can be reformulate as
\begin{equation} \small
    \operatorname{UCT}(v_i^{(l)}) = \frac{Q(v_i^{(l)})}{n_i^{(l)}} + C_1 \sqrt{\frac{\log (n_p^{(l-1)})}{n_i^{(l)}}} + C_2 \cdot G(o_j)^{(l)},
    \label{eq:node-comm:uct-func-with-node-comm}
\end{equation} \normalsize
where $C_2 \in \mathbb{R}_+$ is a hyperparameter controlling the weight of the node communication term. 

\subsection{Hierarchical Node Selection} \label{sec:hierachical_update}

In the search stage, the constructed MCT can be naturally used to find optimal architecture. With the stored Q-value, we can directly sample the architecture with the highest reward as the final result. However, the architecture that performs the best on the training set might not always be the best on the validation set. 
We thus resort to a search stage to evaluate a small pool of the architectures of higher rewards, and then the one with the highest validation accuracy will be exported. Moreover, to accurately evaluate those nodes with small numbers of visits, we propose a hierarchical node selection method to select the node hierarchically and re-evaluate those less-visited nodes.

In MCT, the number of nodes increases exponentially with the depth. It is impossible to visit all the posterior nodes sufficiently during training. Fortunately, as our conditional probability modeling of search space, the sub-trees whose parents have low rewards could be directly trimmed, and we only need to focus on those good sub-trees.

Concretely, as illustrated in the right of Figure \ref{fig:framework}, our selection starts hierarchically from the root node of MCT. If the average number of visits of its child nodes is larger than a threshold constant $n_\mathrm{thrd}$, we think it is promising to its reward, and thus the child node can be sampled using the same probabilistic distribution as Eq.\eqref{eq:softmax-sampling}. On the other hand, if the average number is lower than the $n_\mathrm{thrd}$, we randomly sample paths consisting of those child nodes and then evaluate the paths using a batch of validation data until the threshold reached. Then we have enough confidence to continue moving to the next depth. After selecting the leaf nodes, the specific architecture is obtained, we then evaluate it with the full validation set. 
We repeat this procedure with search number times and then report the one with the highest validation accuracy as the final architecture. Since the ``exploration'' of architectures is not needed in the search stage, the UCT function then becomes: 

\begin{equation}
    \operatorname{UCT}_s(v_i^{(l)}) = \frac{Q(v_i^{(l)})}{n_i^{(l)}} + C_2 \cdot G(o_j).
    \label{eq7}
\end{equation}

Our iterative procedure of searching with hierarchical node selection is shown in Algorithm~\ref{alg:search-stage}.

\begin{algorithm}[tbp]
    \caption{Architecture Search with Hierarchical Node Selection}
    \label{alg:search-stage}
    \KwIn{the root node $v^{(0)}$ of tree $\mathcal{T}$, layer number $L$, search number $K$, validation dataset $\mathcal{D}_{val}$.}
    
    Init $E = \{\}$, $k=0$ ;\\
    \While{$k \le K$}{
        Init $\boldsymbol{\alpha} = \{\}$, $l=0$ ;\\
        \While{$l \leq L$}{
            
            \While{$\frac{1}{N}\sum_{v_i^{(l)} \in \operatorname{child}(v^{(l-1)})} n_i^{(l)} \geq n_{\mathrm{thrd}}$}{
                sample one path $\tilde{\boldsymbol{\alpha}}$ randomly with ancestor nodes $\boldsymbol{\alpha}$; \\
                evaluate $\tilde{\boldsymbol{\alpha}}$ with one batch data from $\mathcal{D}_{val}$;  \\
                update UCT scores corresponding to $\tilde{\boldsymbol{\alpha}}$; \\
            }
            sample a node $v_j^{(l)}$ according to Eq.\eqref{eq:softmax-sampling}; \\
            $\boldsymbol{\alpha} = \boldsymbol{\alpha} \cup \{o_j^{(l)}\}$; \\
        }
        
        evaluate $\boldsymbol{\alpha}$ with validation dataset $\mathcal{D}_{val}$; \\
        $E = E \cup \{\boldsymbol{\alpha}\}$;
    }
    \KwOut{architecture with highest accuracy in $E$}
\end{algorithm}

\section{Experiments}
In this section, we conduct extensive experiments on the proposed NAS benchmark \textit{NAS-Bench-Macro} and ImageNet dataset. Detailed experimental settings are elaborated in supplementary materials.

\subsection{Proposed benchmark: NAS-Bench-Macro} \label{sec:nasbench}
For a better comparison between our MCT-NAS and other methods, we propose an open-source NAS benchmark on macro structures with CIFAR-10 dataset, named NAS-Bench-Macro. The NAS-Bench-Macro consists of $6561$ networks and their test accuracies, parameter numbers, and FLOPs on CIFAR-10.

\textbf{Search space.}  The search space of NAS-Bench-Macro is conducted with 8 searching layers; each layer contains 3 candidate blocks, marked as \textit{Identity}, \textit{MB3\_K3} and \textit{MB6\_K5}. Thus the total size of the search space is $3^8=6561$. However, the architectures with the same \textit{Identity} number in each stage exactly have the same structures and can be mapped together so that the total search space can be mapped to $3969$ architectures. Detailed structure configurations can be found in Supplementary Materials.

\textbf{Benchmarking architectures on CIFAR-10.} We train all architectures isolatedly on CIFAR-10 dataset. Each architecture is trained with a batch size of $256$ and SGD optimizer, a cosine learning rate strategy that decays $50$ epochs is adopted with an initial value $0.1$. We train each architecture $3$ times with different random seeds and report their mean accuracies on the test set.

\textbf{Rank correlations between each search method and NAS-Bench-Macro.} 
In MCT-NAS, during training, the MCTS also participates in the architecture sampling; with UCT function in MCTS, the good architectures can be efficiently explored and trained. Thus, the supernet trained with MCTS should have a more accurate ranking on subnets. We adopt experiments to measure the ranking confidences of these two evaluators trained by uniform sampling and MCTS. Concretely, we calculate the ranking correlation coefficients between the validation accuracies on weight-sharing subnets and their ground-truth performances in NAS-Bench-Macro; the supernets are trained with uniform sampling and MCTS, respectively. The results illustrated in Table~\ref{tab:nasbench_correlation} indicates the supernet trained with our MCTS is more promising to the validation accuracies. However, using MCTS in the whole training stage performs worse than adding uniform sampling for warm-up(\textit{uniform+MCTS}) since the subnets are not converged initially. To adopt an effective exploration in MCTS, we use uniform sampling for warm-up in all our experiments.

\begin{table}
	\centering
	\caption{Rank correlations between each supernet and NAS-Bench-Macro. }  
	\label{tab:nasbench_correlation}
	\begin{tabular}{c||c|c}
		\hline
		training method & Spearman rho (\%) & Kendall tau (\%) \\	
		\hline	
		uniform & 88.96 & 72.41 \\ 
		MCTS & 90.63 & 74.66 \\
		uniform + MCTS & 91.87 & 76.22\\
		\hline
	\end{tabular}
	\vspace{-2mm}
\end{table}

\begin{table*}[t]
	\caption{Comparison of searched architectures \wrt different state-of-the-art NAS methods. Search number means the number of evaluated architectures during searching; our MCT-NAS involves additional evaluation cost in hierarchical node selection method and costs $\sim5\times$ on evaluating one architecture compared to other methods. $\ddagger$: TPU, $\star$: reported by \cite{guo2020single}.} 
	\label{tbl:sota}
	\centering
	\small
 	\vspace{2mm}
	\begin{tabular}{r||cc||cc||cccc} 
		Methods& \tabincell{c}{Top-1\\ (\%)}&\tabincell{c}{Top-5\\ (\%)} & \tabincell{c}{FLOPs\\ (M)} &\tabincell{c}{Params \\ (M)}&\tabincell{c}{Memory cost}&\tabincell{c}{training cost \\ (GPU days)}&\tabincell{c}{search number}&\tabincell{c}{search cost\\ (GPU days)} \\ \hline
		SCARLET-C \cite{chu2019scarletnas}&  75.6&92.6&280 & 6.0 &single path&10& 8400 &12\\ 
		MobileNetV2 1.0 \cite{mobilenetv2} &  72.0& 91.0& 300 &3.4 & - & - & - & - \\
		MnasNet-A1 \cite{mnasnet}& 75.2 & 92.5& 312 &3.9&single path + RL & $288^\ddagger$& 8000 &-\\
		GreedyNAS-C \cite{you2020greedynas} & 76.2 & 92.5 & 284 & 4.7 & single path & 7 & 1000 & $<1$ \\
		MCT-NAS-C &\textbf{76.3}&92.6&280& 4.9&single path&12& 20$\times$5&$<1$\\ \hline	
		Proxyless-R (mobile) \cite{proxylessnas}&74.6&92.2&320&4.0&two paths&$15^\star$& 1000 &-\\
		Single-path \cite{guo2020single}& 76.2&-&328& - &single path&12&1000&$<1$\\
		ST-NAS-A \cite{powering} &  76.4 & 93.1 & 326 & 5.2 & single path & - & 990 & -  \\
		SCARLET-B \cite{chu2019scarletnas}&  76.3&93.0&329& 6.5 &single path&10&8400&12\\ 
		GreedyNAS-B \cite{you2020greedynas} & 76.8 & 93.0 & 324 & 5.2 & single path & 7 &1000& $<1$ \\
		FairNAS-C \cite{fairnas}&  76.7&93.3&325& 5.6 &single path&-&-\\ 
		BetaNet-A \cite{fang2019betanas}& 75.9 & 92.8 & 333 & 4.1 &single path& 7 & - & - \\
		MCT-NAS-B&\textbf{76.9} &93.4&327 &6.3 &single path&12& $20\times 5$&$<1$\\ \hline\  
		ST-NAS-B \cite{powering} &  77.9 & 93.8 & 503 & 7.8 & single path & - & 990 & -  \\
		BetaNet-A $\times$ 1.4 \cite{fang2019betanas}& 77.7 & 93.7 & 631 & 7.2 &single path& 7 & - & - \\
		EfficientNet-B0 \cite{tan2019efficientnet} & 76.3 & 93.2& 390 & 5.3 &single path& - & - & - \\ 
		MCT-NAS-A&\textbf{78.0} &93.9&442 &8.4 &single path&12& $20\times 5$&$<1$\\ \hline
	\end{tabular} 
	\label{nas_experiments}
	\vspace{-4mm}
\end{table*}

\begin{figure}
\centering
    \subfigure[Top ACCs]
   {\includegraphics[width=0.85\linewidth]{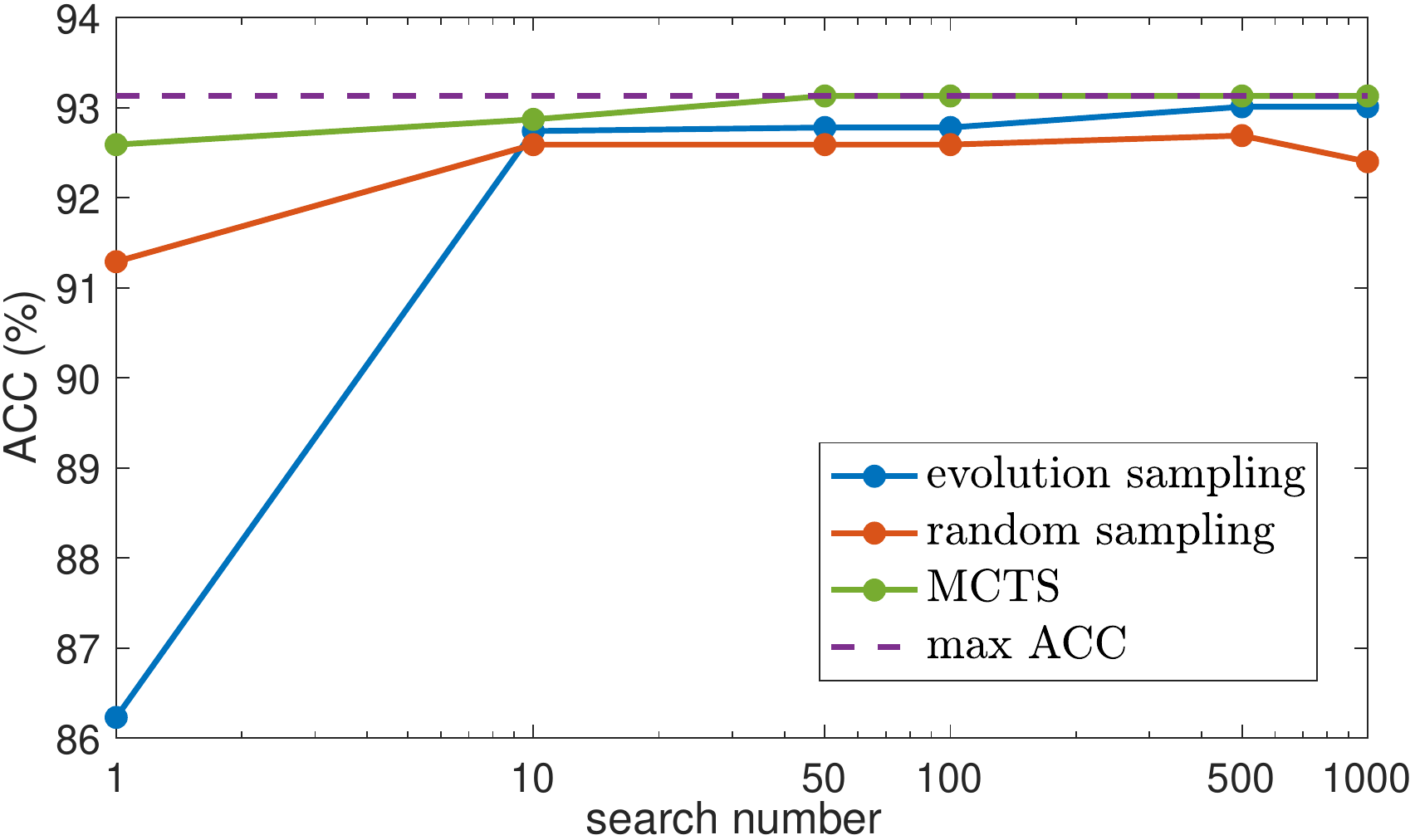}}
   \subfigure[Average Percentile Rank]
   {\includegraphics[width=0.85\linewidth]{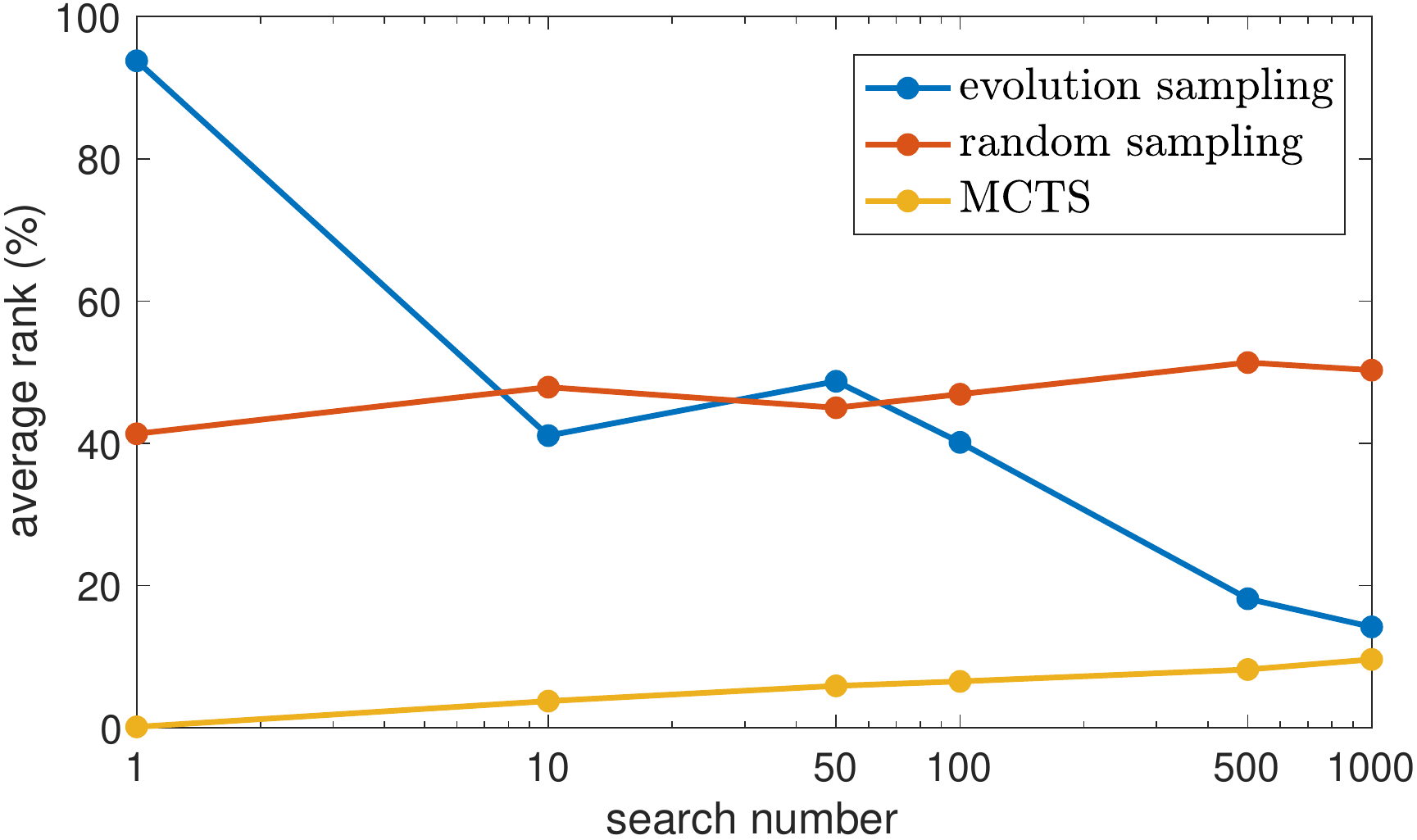}}
  \caption{Top accuracies (a) and average percentile rank (b) of searched architectures with different search numbers in NAS-Bench-Macro.}
\label{fig:nasbench_search_number}
  \vspace{-2mm}
\end{figure}

\textbf{Comparison between different search numbers.}
Before searching, the MCT has already stored each subnet's reward, so we can directly use the subnet with the highest reward as the final architecture. However, these rewards are restricted to the training loss, thus not accurate enough to the performances, so the subnets still need to be evaluated on the validation set for better results. We adopt experiments with different search numbers and report the top accuracy and average percentile rank with NAS-Bench-Macro. The result illustrated in Figure \ref{fig:nasbench_search_number} shows that, by directly using the max-rewarded architecture in MCT, our MCT-NAS can also obtain higher performance compared to random search and evolutionary algorithm. However, more explorations obtain even better results, and only with a search number of $50$, our MCT-NAS can find the best architecture in NAS-Bench-Macro. 
Meanwhile, as Figure \ref{fig:nasbench_search_number} (b) shows, the lower average percentile rank indicates the better average performance of networks. Since our MCT-NAS stores an accurate ranking of architectures, good architectures are usually explored in the early stage; with the increment of search number, the average performance will drop. However, the evolution shows a reverse trend since it starts with a random population and utilizes the evaluation results for the new population. According to the results, we set the search number to $20$ in all experiments to a better trade-off between performance and search efficiency.

\subsection{Search on ImageNet}
\textbf{Dataset.} 
We conduct the architecture search on the large-scale dataset ImageNet (ILSVRC-12), consisting of 1.28 million training images and 50k validation images from 1000 categories. To facilitate the search, 
we randomly sample 50k images from the training dataset as the validation dataset, and the rest of the data is used for training. For comparison with other methods, we use the original validation dataset as a test dataset to report the accuracy.

\textbf{Search space.} For a fair comparison, we conduct our MCTS-NAS  with the same space as \cite{you2020greedynas,chu2019scarletnas,proxylessnas} to examine the performance of other one-shot NAS methods with macro search space. We construct the supernet with $21$ search blocks, and each block is a MobileNetV2 inverted bottleneck \cite{mobilenetv2} with an optional SE \cite{hu2018squeeze} module. For each search block, the convolutional kernel size is within $\{3,5,7\}$, and the expansion ratio is selected in $\{3,6\}$, and each block can choose to use SE module or not. With an additional identity block for network depth search, the total operation space size is $13$, so the search space size is $13^{21}$.

\textbf{Supernet training.} Follow \cite{you2020greedynas, guo2020single}, we use the same strategy for training supernet. Using a batch size of $1024$, the network is trained using a SGD optimizer with $0.9$ momentum. A cosine annealing strategy is adopted with an initial learning rate $0.12$, which decays 120 epochs. For sampling architectures, we use uniform sampling within the first 60 epochs, then adopt our prioritized sampling with MCT, and the temperature term $\tau$ is set to $0.0025$ in all of our experiments for appropriate sampling paths. We conduct experiments on three different FLOPs budgets $280$M, $330$M, and $440$M.
Note that for a better adaptation to the target FLOPs budgets, we only sample the architectures within the target FLOPs, this reduction of search space results in a better convergence and evaluation of potential architectures. 

\textbf{Searching. } We use hierarchical selection with MCT for the search, which is summarized in Algorithm \ref{alg:search-stage}. The number of search architectures is set to $20$ for efficiency.

\textbf{Retraining. } To train the obtain architectures from scratch, we follow previous works \cite{mnasnet, you2020greedynas, chu2019scarletnas}, the network is trained using RMSProp optimizer with 0.9 momentum, and the learning rate is increased from 0 to $0.064$ linearly in the first $5$ epochs with batch size 512, and then decays 0.03 every 2.4 epochs. Besides, the exponential moving average on weights is also adopted with a decay rate 0.9999.

\textbf{Performances of obtained architectures.}  We perform the search with 3 different FLOPs budgets, \ie, 442M, 327M, and 280M. As shown in Table \ref{nas_experiments}, our searched 442M MCT-NAS-A achieves 78.0\% on Top-1 accuracy, which even outperforms other methods with larger FLOPs budget (\ie, 503M and 631M). Besides, with other FLOPs (\ie, 327M and 280M), our MCT-NAS also shows superiority over other methods. Besides, MCT-NAS also achieves superiority in terms of searching efficiency, as in Table \ref{nas_experiments}, we perform the search with only 20 sampled paths with our proposed MCT-NAS, which amounts to about one-tenth of the paths of other algorithms.

\begin{table*}[t]
    \renewcommand\arraystretch{1.17}
	\setlength\tabcolsep{1.5mm}
	\caption{The performance gain of each part in MCT-NAS with 330M FLOPs on MobileNet search space.}
	\label{MCT_NAS_analysis}
	\centering
	\scriptsize
	\begin{tabular}{|c|c|c|c|c|c|c|c|c||c|} \hline
		\multicolumn{5}{|c|}{Training} & \multicolumn{3}{c|}{Searching}&  \multicolumn{2}{c|}{Retraining} \\ \cline{1-10} 
		Uniform & FLOPs & Update MCT & UCT & Node & \multicolumn{1}{c|}{Evolutionary}& \multicolumn{1}{c}{MCTS} & Hierarchical & \multirow{2}*{Top-1} & \multirow{2}*{Top-5}\\ 
		Sampling &Reduction&in Training& Search & Communication &Search & Search  &  Updates &  &  \\ \hline 
		\checkmark& & & & & \checkmark & &  & 75.94\% & 92.89\% \\
		\checkmark& & & & &  & \checkmark & \checkmark & 76.44\% & 93.15\% \\
		\checkmark & \checkmark &\checkmark& & & & \checkmark & & 76.21\% & 93.11\%  \\
		\checkmark & \checkmark &\checkmark& \checkmark & & & \checkmark & & 76.35\% & 93.17\% \\ 
		\checkmark& \checkmark & \checkmark & \checkmark & \checkmark & &  \checkmark & & 76.62\% & 93.32\% \\
		\checkmark& \checkmark & \checkmark & \checkmark & \checkmark & &  \checkmark & \checkmark & 76.94\% & 93.37\% \\ \hline
	\end{tabular} 
	\vspace{-2mm}
\end{table*}

\subsection{Ablation Studies}

\textbf{Ablation on each proposed techniques in MCT-NAS}.  In MCT-NAS, as summarized in Table~\ref{MCT_NAS_analysis}, at the supernet training stage, we propose to update MCT using the training loss, for sampling subnets, a UCT search strategy is adopted. Besides, we further propose a node communication technique to make use of the rewards of operations. While on search, we propose a hierarchical node selection method for a more accurate search using MCT. We evaluate the performances with different combinations of these techniques on ImageNet and report the test accuracies in Table~\ref{MCT_NAS_analysis}.  FLOPs' reduction indicates removing the architectures in MCT whose FLOPs are far smaller or bigger than the target FLOPs.
Comparing row 3 and row 4, we can infer that, with UCT search, the obtained architecture achieves higher performance. That might be because sampling subnets with UCT makes a better exploration-exploitation balance and focuses more on the good subnets than uniform sampling. 
Comparing rows 1, 2, and 3, using MCTS during search obtains better results compared to the evolution search since extensive information in the training stage is used.
For row 5 and row 6, the architecture with hierarchical selection is significantly better because the hierarchical selection adopts additional evaluations using validation data.
With all the proposed methods equipped, our MCT-NAS achieves the highest performance.

\begin{figure}[t]
\begin{center}
\includegraphics[width=0.9\linewidth]{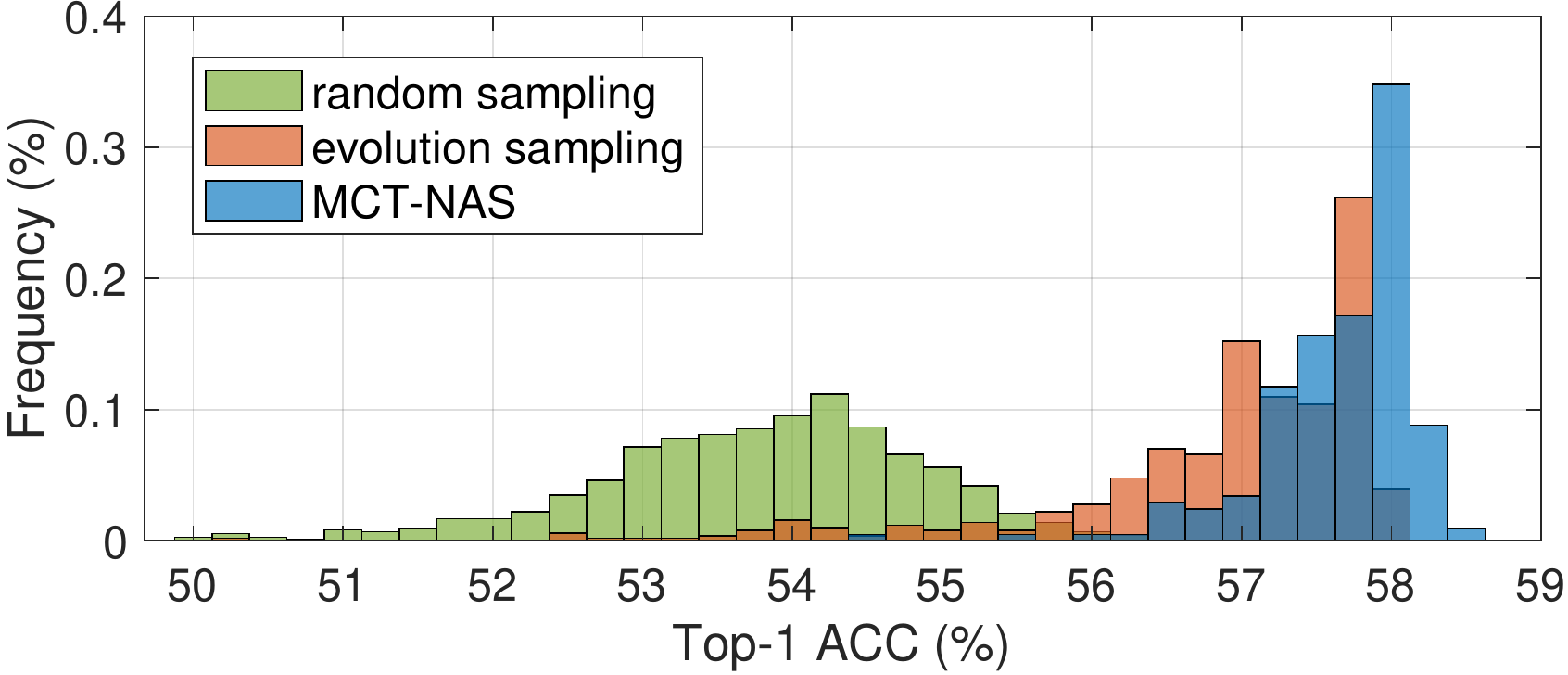}
\end{center}
    \vspace{-3mm}
  \caption{Histogram of accuracy of searched architectures on supernet with different search methods.}
  \vspace{-4mm}
\label{histogram}
\end{figure}

\textbf{Comparison of different sampling methods.} To examine the searching efficiency and results \wrt different search methods, we searching for 330M-FLOPs paths by evolution sampling, random sampling, and MCT-NAS, respectively, using the same supernet trained by uniform sampling on ImageNet. In this way, we examine the Top-1 accuracy \wrt the 1000 search numbers with different search methods and show their histogram in Figure \ref{histogram}. In detail, MCT-NAS and evolution sampling both works as heuristic searching methods with prior knowledge, which can search for subnets with much higher Top-1 ACC than random sampling. Besides, as in Figure \ref{histogram}, the paths searched by our MCT-NAS can be more concentrated in areas with higher Top-1 accuracy, while for evolution sampling, the performance of searched subnets is distributed in a larger area, which contains many paths with sub-optimal results. While for MCT-NAS, 
the performance of searched paths is mostly located at the area with good performance, which indicates the effectiveness of our sampling method.

\begin{figure}[t]
\begin{center}
\includegraphics[width=0.95\linewidth]{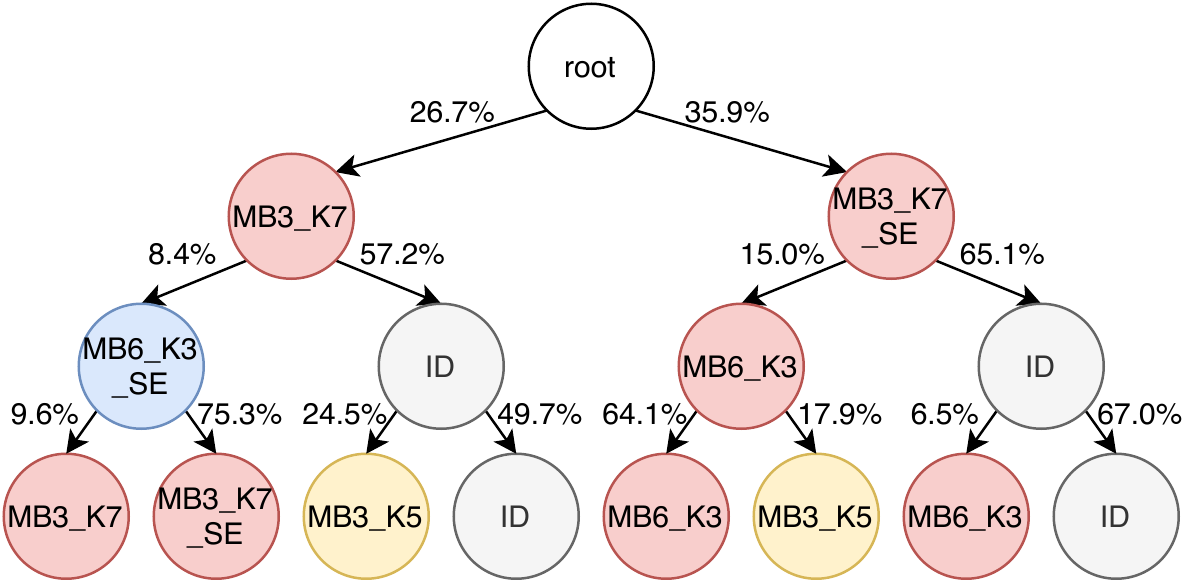}
\end{center}
\vspace{-3mm}
  \caption{Interpretation and visualization of nodes selection. The marked texts in circles denote candidate operations, which will be introduced in supplementary materials. }
 \vspace{-6mm}
\label{node_visual}
\end{figure}

\textbf{Interpretation and visualization of nodes selection.} 
As a tree-based structure, the MCT obtained in our method captures the dependencies between different layers. We believe that the operation choice of one layer is related to its previous layers, \ie, different choices of previous layers may result in different preferences of the layer. For an intuitive understanding, we visualize the nodes with top-2 UCT scores of the first 3 layers in Figure \ref{node_visual}. The visualization shows that when the first layer selects different nodes (\ie, MB3\_K7 and MB3\_K7\_SE), the nodes with the Top-2 selection probability of the next layer are greatly affected. As a result, MCTS works as a path-level subnet sampling method that naturally captures the dependencies between operations of different layers and selects paths with global relations more efficiently. Nevertheless, other sampling methods are usually implemented at the node level, which selects nodes separately and ignores the dependencies between nodes, thus leading to sub-optimal results.

\section{Conclusion}
In this paper, we introduce a sampling strategy based on Monte Carlo tree search (MCTS), which models the search space as a Monte Carlo tree (MCT), and naturally captures the dependencies between layers. Furthermore, with the extensive intermediate results updated in MCT during training, we propose node communication technique and hierarchical selection of MCT to make better use of the information, and thus the optimal architecture can be efficiently obtained. To better compare different NAS methods, we construct a NAS benchmark on macro search space with CIFAR-10, named NAS-Bench-Macro. Extensive experiments on NAS-Bench-Macro and ImageNet demonstrate that our MCT-NAS significantly improves search efficiency and performance.

\subsubsection*{Acknowledgments}
This work is funded by the National Key Research and Development Program of China (No. 2018AAA0100701) and the NSFC 61876095. Chang Xu was supported in part by the Australian Research Council under Projects DE180101438 and DP210101859.

{\small
\bibliographystyle{ieee_fullname}
\bibliography{egbib}

\begin{thebibliography}{10}\itemsep=-1pt

\bibitem{proxylessnas}
Han Cai, Ligeng Zhu, and Song Han.
\newblock Proxylessnas: Direct neural architecture search on target task and
  hardware.
\newblock {\em arXiv preprint arXiv:1812.00332}, 2018.

\bibitem{chu2019scarletnas}
Xiangxiang Chu, Bo Zhang, Jixiang Li, Qingyuan Li, and Ruijun Xu.
\newblock Scarletnas: Bridging the gap between scalability and fairness in
  neural architecture search.
\newblock {\em arXiv preprint arXiv:1908.06022}, 2019.

\bibitem{fairnas}
Xiangxiang Chu, Bo Zhang, Ruijun Xu, and Jixiang Li.
\newblock Fairnas: Rethinking evaluation fairness of weight sharing neural
  architecture search.
\newblock {\em arXiv preprint arXiv:1907.01845}, 2019.

\bibitem{deb2002fast}
Kalyanmoy Deb, Amrit Pratap, Sameer Agarwal, and TAMT Meyarivan.
\newblock A fast and elitist multiobjective genetic algorithm: Nsga-ii.
\newblock {\em IEEE transactions on evolutionary computation}, 6(2):182--197,
  2002.

\bibitem{du2020agree}
Shangchen Du, Shan You, Xiaojie Li, Jianlong Wu, Fei Wang, Chen Qian, and
  Changshui Zhang.
\newblock Agree to disagree: Adaptive ensemble knowledge distillation in
  gradient space.
\newblock {\em Advances in Neural Information Processing Systems}, 33, 2020.

\bibitem{fang2019betanas}
Muyuan Fang, Qiang Wang, and Zhao Zhong.
\newblock Betanas: Balanced training and selective drop for neural architecture
  search.
\newblock {\em arXiv preprint arXiv:1912.11191}, 2019.

\bibitem{guo2020hit}
Jianyuan Guo, Kai Han, Yunhe Wang, Chao Zhang, Zhaohui Yang, Han Wu, Xinghao
  Chen, and Chang Xu.
\newblock Hit-detector: Hierarchical trinity architecture search for object
  detection.
\newblock In {\em Proceedings of the IEEE/CVF Conference on Computer Vision and
  Pattern Recognition}, pages 11405--11414, 2020.

\bibitem{powering}
Ronghao Guo, Chen Lin, Chuming Li, Keyu Tian, Ming Sun, Lu Sheng, and Junjie
  Yan.
\newblock Powering one-shot topological nas with stabilized share-parameter
  proxy.
\newblock {\em arXiv preprint arXiv:2005.10511}, 2020.

\bibitem{guo2020single}
Zichao Guo, Xiangyu Zhang, Haoyuan Mu, Wen Heng, Zechun Liu, Yichen Wei, and
  Jian Sun.
\newblock Single path one-shot neural architecture search with uniform
  sampling.
\newblock In {\em European Conference on Computer Vision}, pages 544--560.
  Springer, 2020.

\bibitem{han2020ghostnet}
Kai Han, Yunhe Wang, Qi Tian, Jianyuan Guo, Chunjing Xu, and Chang Xu.
\newblock Ghostnet: More features from cheap operations.
\newblock In {\em Proceedings of the IEEE/CVF Conference on Computer Vision and
  Pattern Recognition}, pages 1580--1589, 2020.

\bibitem{han2021transformer}
Kai Han, An Xiao, Enhua Wu, Jianyuan Guo, Chunjing Xu, and Yunhe Wang.
\newblock Transformer in transformer.
\newblock {\em arXiv preprint arXiv:2103.00112}, 2021.

\bibitem{hinton2015distilling}
Geoffrey Hinton, Oriol Vinyals, and Jeff Dean.
\newblock Distilling the knowledge in a neural network.
\newblock {\em arXiv preprint arXiv:1503.02531}, 2015.

\bibitem{hu2018squeeze}
Jie Hu, Li Shen, and Gang Sun.
\newblock Squeeze-and-excitation networks.
\newblock In {\em Proceedings of the IEEE conference on computer vision and
  pattern recognition}, pages 7132--7141, 2018.

\bibitem{huang2020explicitly}
Tao Huang, Shan You, Yibo Yang, Zhuozhuo Tu, Fei Wang, Chen Qian, and Changshui
  Zhang.
\newblock Explicitly learning topology for differentiable neural architecture
  search.
\newblock {\em arXiv preprint arXiv:2011.09300}, 2020.

\bibitem{kendall1938new}
Maurice~G Kendall.
\newblock A new measure of rank correlation.
\newblock {\em Biometrika}, 30(1/2):81--93, 1938.

\bibitem{kocsis2006bandit}
Levente Kocsis and Csaba Szepesv{\'a}ri.
\newblock Bandit based monte-carlo planning.
\newblock In {\em European conference on machine learning}, pages 282--293.
  Springer, 2006.

\bibitem{li2020sgas}
Guohao Li, Guocheng Qian, Itzel~C Delgadillo, Matthias Muller, Ali Thabet, and
  Bernard Ghanem.
\newblock Sgas: Sequential greedy architecture search.
\newblock In {\em Proceedings of the IEEE/CVF Conference on Computer Vision and
  Pattern Recognition}, pages 1620--1630, 2020.

\bibitem{Liao_2020_CVPR}
Yue Liao, Si Liu, Fei Wang, Yanjie Chen, Chen Qian, and Jiashi Feng.
\newblock Ppdm: Parallel point detection and matching for real-time
  human-object interaction detection.
\newblock In {\em Proceedings of the IEEE/CVF Conference on Computer Vision and
  Pattern Recognition (CVPR)}, June 2020.

\bibitem{liu2018darts}
Hanxiao Liu, Karen Simonyan, and Yiming Yang.
\newblock Darts: Differentiable architecture search.
\newblock {\em arXiv preprint arXiv:1806.09055}, 2018.

\bibitem{negrinho2017deeparchitect}
Renato Negrinho and Geoff Gordon.
\newblock Deeparchitect: Automatically designing and training deep
  architectures.
\newblock {\em arXiv preprint arXiv:1704.08792}, 2017.

\bibitem{pirie2004spearman}
W Pirie.
\newblock Spearman rank correlation coefficient.
\newblock {\em Encyclopedia of statistical sciences}, 12, 2004.

\bibitem{russakovsky2015imagenet}
Olga Russakovsky, Jia Deng, Hao Su, Jonathan Krause, Sanjeev Satheesh, Sean Ma,
  Zhiheng Huang, Andrej Karpathy, Aditya Khosla, Michael Bernstein,
  Alexander~C. Berg, and Li Fei-Fei.
\newblock {ImageNet Large Scale Visual Recognition Challenge}.
\newblock {\em International Journal of Computer Vision (IJCV)},
  115(3):211--252, 2015.

\bibitem{mobilenetv2}
Mark Sandler, Andrew Howard, Menglong Zhu, Andrey Zhmoginov, and Liang-Chieh
  Chen.
\newblock Mobilenetv2: Inverted residuals and linear bottlenecks.
\newblock In {\em Proceedings of the IEEE conference on computer vision and
  pattern recognition}, pages 4510--4520, 2018.

\bibitem{su2021locally}
Xiu Su, Shan You, Tao Huang, Fei Wang, Chen Qian, Changshui Zhang, and Chang
  Xu.
\newblock Locally free weight sharing for network width search.
\newblock {\em arXiv preprint arXiv:2102.05258}, 2021.

\bibitem{mnasnet}
Mingxing Tan, Bo Chen, Ruoming Pang, Vijay Vasudevan, Mark Sandler, Andrew
  Howard, and Quoc~V Le.
\newblock Mnasnet: Platform-aware neural architecture search for mobile.
\newblock In {\em Proceedings of the IEEE Conference on Computer Vision and
  Pattern Recognition}, pages 2820--2828, 2019.

\bibitem{tan2019efficientnet}
Mingxing Tan and Quoc~V Le.
\newblock Efficientnet: Rethinking model scaling for convolutional neural
  networks.
\newblock {\em arXiv preprint arXiv:1905.11946}, 2019.

\bibitem{tang2020semi}
Yehui Tang, Yunhe Wang, Yixing Xu, Hanting Chen, Boxin Shi, Chao Xu, Chunjing
  Xu, Qi Tian, and Chang Xu.
\newblock A semi-supervised assessor of neural architectures.
\newblock In {\em Proceedings of the IEEE/CVF Conference on Computer Vision and
  Pattern Recognition}, pages 1810--1819, 2020.

\bibitem{tang2020reborn}
Yehui Tang, Shan You, Chang Xu, Jin Han, Chen Qian, Boxin Shi, Chao Xu, and
  Changshui Zhang.
\newblock Reborn filters: Pruning convolutional neural networks with limited
  data.
\newblock In {\em AAAI}, pages 5972--5980, 2020.

\bibitem{wang2020learning}
Linnan Wang, Rodrigo Fonseca, and Yuandong Tian.
\newblock Learning search space partition for black-box optimization using
  monte carlo tree search.
\newblock {\em Advances in Neural Information Processing Systems}, 33, 2020.

\bibitem{wang2019sample}
Linnan Wang, Saining Xie, Teng Li, Rodrigo Fonseca, and Yuandong Tian.
\newblock Sample-efficient neural architecture search by learning action space.
\newblock {\em arXiv preprint arXiv:1906.06832}, 2019.

\bibitem{wang2020neural}
Linnan Wang, Yiyang Zhao, Yuu Jinnai, Yuandong Tian, and Rodrigo Fonseca.
\newblock Neural architecture search using deep neural networks and monte carlo
  tree search.
\newblock In {\em Proceedings of the AAAI Conference on Artificial
  Intelligence}, volume~34, pages 9983--9991, 2020.

\bibitem{wei2020point}
Fangyun Wei, Xiao Sun, Hongyang Li, Jingdong Wang, and Stephen Lin.
\newblock Point-set anchors for object detection, instance segmentation and
  pose estimation.
\newblock In {\em European Conference on Computer Vision}, pages 527--544.
  Springer, 2020.

\bibitem{yang2020ista}
Yibo Yang, Hongyang Li, Shan You, Fei Wang, Chen Qian, and Zhouchen Lin.
\newblock Ista-nas: Efficient and consistent neural architecture search by
  sparse coding.
\newblock {\em arXiv preprint arXiv:2010.06176}, 2020.

\bibitem{yang2021towards}
Yibo Yang, Shan You, Hongyang Li, Fei Wang, Chen Qian, and Zhouchen Lin.
\newblock Towards improving the consistency, efficiency, and flexibility of
  differentiable neural architecture search.
\newblock {\em arXiv preprint arXiv:2101.11342}, 2021.

\bibitem{you2020greedynas}
Shan You, Tao Huang, Mingmin Yang, Fei Wang, Chen Qian, and Changshui Zhang.
\newblock Greedynas: Towards fast one-shot nas with greedy supernet.
\newblock In {\em Proceedings of the IEEE/CVF Conference on Computer Vision and
  Pattern Recognition}, pages 1999--2008, 2020.

\bibitem{you2017learning}
Shan You, Chang Xu, Chao Xu, and Dacheng Tao.
\newblock Learning from multiple teacher networks.
\newblock In {\em Proceedings of the 23rd ACM SIGKDD International Conference
  on Knowledge Discovery and Data Mining}, pages 1285--1294, 2017.

\bibitem{zhou2019bayesnas}
Hongpeng Zhou, Minghao Yang, Jun Wang, and Wei Pan.
\newblock Bayesnas: A bayesian approach for neural architecture search.
\newblock {\em arXiv preprint arXiv:1905.04919}, 2019.

\end{thebibliography}
}

\onecolumn
\appendix

\section{Detailed Experimental Settings}
In this section, we will describe the detailed experimental settings of our MCT-NAS \wrt training and searching. In general, we used the same MCT during training and searching, and thus the constructed MCT with prioritized sampled paths can boost the search in terms of efficiency and search performance.

\textbf{Supernet training with MCT-NAS.} For ImageNet dataset, we use the same strategy follow \cite{you2020greedynas, guo2020single}. In detail, we use a batch size of 1024, and train the supernet using a SGD optimizer with 0.9 momentum. A cosine annealing strategy is adopted with an initial learning rate 0.12, which decays 120 epochs. For sampling architectures, we first warm up the supernet with 50\% of overall training epochs (60 epochs) with uniform sampling. Then for the last 60 epochs, we sample subnets with FLOPs reduction, \ie, the FLOPs of sampled subnets must be within a certain range of FLOPs budget(\ie, 0.9$\times$ $\sim$ 1.0$\times$).
In the 20\% of following training epochs (61-85 epochs), we also uniformly sample subnets to construct the MCT. Afterward, in the last 34 epochs (86-120), we sample subnets with MCT and UCT function defined in Eq. \eqref{eq:softmax-sampling} and Eq. \eqref{eq:node-comm:uct-func-with-node-comm}. In detail, we adopt $C_1=0.1$ and $C_2=0.2$ for Eq. \eqref{eq:node-comm:uct-func-with-node-comm} and Eq. \eqref{eq7}.
Besides, for NAS-Bench-Macro, we follow the same split ratios of using uniform sampling and MCTS as on ImageNet. And other training strategies are the same as retraining on CIFAR-10.

\textbf{Searching optimal structure with MCT-NAS.} We use hierarchical node selection with MCT for architecture search. For sampling a path (subnet), we select the optimal nodes hierarchically from the root node of MCT with a threshold constant $n_{thrd}$ of $6$. If the average number of visits of its child nodes is lower than the $n_{thrd}$, we randomly sample paths consisting of those child nodes and then evaluate the paths using a batch (\ie, 128) of validation data until the threshold reached. After selecting the leaf nodes, the specific subnet (structure) is obtained; we then evaluate it with the full validation dataset. Moreover, we repeat to sample subnets with this process until we reach our predefined number of searches (\ie, 20). Afterward, we select the structure with the best validation accuracy to train from scratch for evaluation.

\textbf{Retraining of searched optimal structures.} To train the obtained architectures from scratch, we follow previous works \cite{mnasnet, you2020greedynas, chu2019scarletnas}, the network is trained using RMSProp optimizer with 0.9 momentum, and the learning rate is increased from 0 to $0.064$ linearly in the first $5$ epochs with batch size 512, and then decays 0.03 every 2.4 epochs. Besides, the exponential moving average on weights is also adopted with a decay rate 0.9999.

\section{Details of Path Sampling in Supernet Training}

    The supernet training in our framework can be divided into three stages with different subnet sampling strategies, including the supernet warm-up, the MCT warm-up, and sampling with MCTS.
    Firstly, in the supernet warm-up stage, we sample each path uniformly for a better search space exploration, which is the same as \cite{guo2020single, chu2019scarletnas}.
    Then, in the MCT warm-up stage, we also sample each path uniformly but discard those who do not meet the computation budget (\ie, a certain range of FLOPs budget). Architectures with low computation budget are expected to have lower performance, which makes them less worthy of being optimized and evaluated. In this way, we can construct the MCT more efficiently.
    Finally, we sample paths according to the UCT function in Eq. \eqref{eq:node-comm:uct-func-with-node-comm} to train the supernet more efficiently with exploration-exploitation strategy in MCTS.
    Our iterative procedure of training supernet is presented in Algorithm \ref{supp_algorithm2}.

\newpage

\begin{algorithm}[H]
  \label{supp_algorithm2}
  \SetAlgoLined
  \KwIn{the ratio of iterations for supernet warm-up $W_S$, the ratio of iterations for MCT warm-up $W_M$, training dataset $\mathcal{D}_{tr}$ the maximum training iterations $N$, FLOPs budget $\mathcal{B}$.}

  \While{training epochs $<N$}{
    \uIf{training epochs $<W_S\times N$}{
        randomly sample a path $p$ in the supernet\;
        optimize the path $p$ with dataset $\mathcal{D}_{tr}$;
        }
        
    \Else{
        \uIf{training epochs $<(W_S+W_M)\times N$}{
        random sample a path $p$\;
        \While{FLOPs of path $p$ not in the certain range of $\mathcal{B}$}{           re-sample a new path $p$\;
            }
        
        }
        \Else{
        sample a path $p$ based on MCT with  Hierarchical Node Selection as Algorithm \ref{alg:search-stage}  \;
        }
        optimize the path $p$ with dataset $\mathcal{D}_{tr}$\; 
        updating the nodes in MCT which corresponds to path $p$ with the training loss\;
    }
    
}

  \caption{Path Sampling in Supernet Training}
\end{algorithm}

\section{Details  of Search Space on ImageNet Dataset}
    
    The macros-structure of the supernet used in our experiments on ImageNet is as described in Table \ref{tab:appendix:search-space-imagenet}. The choices for building blocks are as listed in Table \ref{tab:appendix:choice-block}.
    
    \begin{table}[!htbp]
        \centering
        \caption{The macro-structure of the supernet in our experiments on ImageNet.
        The mean of each column is as follow: "n" is the number of such stacked building blocks; "input" is the size of input feature map; "block" is the type of building block; "channels" is the number of output channels, i.e. the number of filters; and "stride" is the stride of the first block among several repeated building blocks. The block type "Choice Block" means selecting a candidate from Table \ref{tab:appendix:choice-block}.}
        \label{tab:appendix:search-space-imagenet}
        \begin{tabular}{c|c|c|c|c}
            \hline
                n & input & block & channels & stride \\
            \hline
                1 & $224 \times 224 \times    3$ & $3 \times 3$ conv & $  32$ & 2 \\
                1 & $112 \times 112 \times   32$ & MB1\_K3           & $  16$ & 1 \\
                4 & $112 \times 112 \times   16$ & Choice Block      & $  32$ & 2 \\
                4 & $ 56 \times  56 \times   32$ & Choice Block      & $  40$ & 2 \\
                4 & $ 28 \times  28 \times   40$ & Choice Block      & $  80$ & 2 \\
                4 & $ 14 \times  14 \times   80$ & Choice Block      & $  96$ & 1 \\
                4 & $ 14 \times  14 \times   96$ & Choice Block      & $ 192$ & 2 \\
                1 & $  7 \times   7 \times  192$ & Choice Block      & $ 320$ & 1 \\
                1 & $  7 \times   7 \times  320$ & $1 \times 1$ conv & $1280$ & 1 \\
                1 & $  7 \times   7 \times 1280$ & global avgpool    & -      & - \\
                1 & 1280                         & FC                & $1000$ & - \\
            \hline
        \end{tabular}
    \end{table}
    
    \begin{table}[!htbp]
        \centering
        \caption{The operation candidates for the MobileNetV2-like building blocks used in Table \ref{tab:appendix:search-space-imagenet}. "ID" denotes an identity mapping.}
        \label{tab:appendix:choice-block}
        \begin{tabular}{c|c|c|c}
            \hline
                block type & expansion ratio & kernel & SE \\
            \hline
                MB1\_K3     & 1 & 3 & no  \\
            \hline
                ID          & - & - & -   \\
                MB3\_K3     & 3 & 3 & no  \\
                MB3\_K5     & 3 & 5 & no  \\
                MB3\_K7     & 3 & 7 & no  \\
                MB6\_K3     & 6 & 3 & no  \\
                MB6\_K5     & 6 & 5 & no  \\
                MB6\_K7     & 6 & 7 & no  \\
            \hline
                MB3\_K3\_SE & 3 & 3 & yes \\
                MB3\_K5\_SE & 3 & 5 & yes \\
                MB3\_K7\_SE & 3 & 7 & yes \\
                MB6\_K3\_SE & 6 & 3 & yes \\
                MB6\_K5\_SE & 6 & 5 & yes \\
                MB6\_K7\_SE & 6 & 7 & yes \\
            \hline
        \end{tabular}
    \end{table}

\section{Details of Proposed Benchmark: NAS-Bench-Macro}

\textbf{Meaning of NAS-Bench-Macro.} The key challenge of the one-shot NAS algorithm lies in the evaluation and ranking reliability of the supernet, which can be reflected by the ranking correlation between the evaluation performances of all architectures on supernet and their actual performances. There are already several NAS benchmarks for the DARTS-like micro search space. However, currently, there is no (as far as we know) public benchmark for one-shot macro search space that provides performances of all networks by retraining from scratch. Therefore, we construct NAS-Bench-Macro with the controlled size of the search space for the toy experiment of verifying our method's effectiveness in terms of searching efficiency and performance. Indeed, we construct NAS-Bench-Macro with the knowledge from MobileNetV2 search space but remove 7$\times$7 convolution for simplifying.

\subsection{Details of search space on NAS-Bench-Macro}
The macro structure of our search space is presented in Table \ref{tab:nasbench_search_space}.
\begin{table}[h]
    \centering
    \caption{Macro structure of search space on NAS-Bench-Macro.}
    \begin{tabular}{c|c|c|c|c}
        \hline
        n & input & block & channel & stride \\
        \hline
        1 & $32\times32\times3$ & $3\times3$ conv&  32 & 1\\
        2 & $32\times32\times32$ & Choice Block & 64 & 2 \\
        3 & $16\times16\times64$ & Choice Block & 128 & 2 \\
        3 & $8\times8\times128$ & Choice Block & 256 & 2 \\
        1 & $4\times4\times256$ & $1\times1$ conv & 1280 & 1 \\
        1 & $4\times4\times1280$ & global avgpool & - & - \\
        1 & $1280$ & FC & 10 & - \\
        \hline
    \end{tabular}
    
    \label{tab:nasbench_search_space}
\end{table}

\subsection{Statistics on CIFAR-10 training results}
\begin{figure}[h]
    \centering
    \subfigure[Params]{\includegraphics[width=0.3\linewidth]{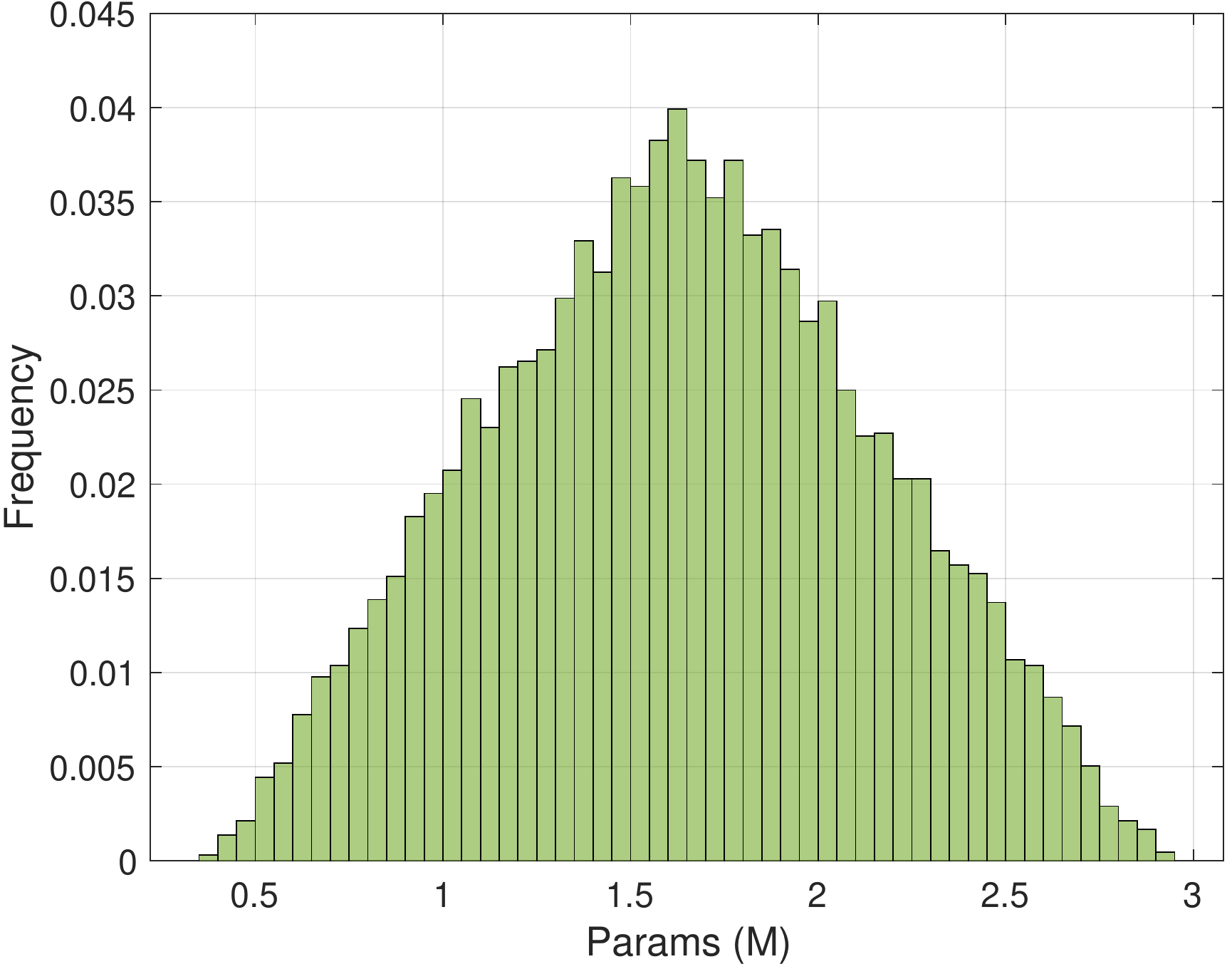}}
    \hspace{2mm}
    \subfigure[FLOPs]{\includegraphics[width=0.3\linewidth]{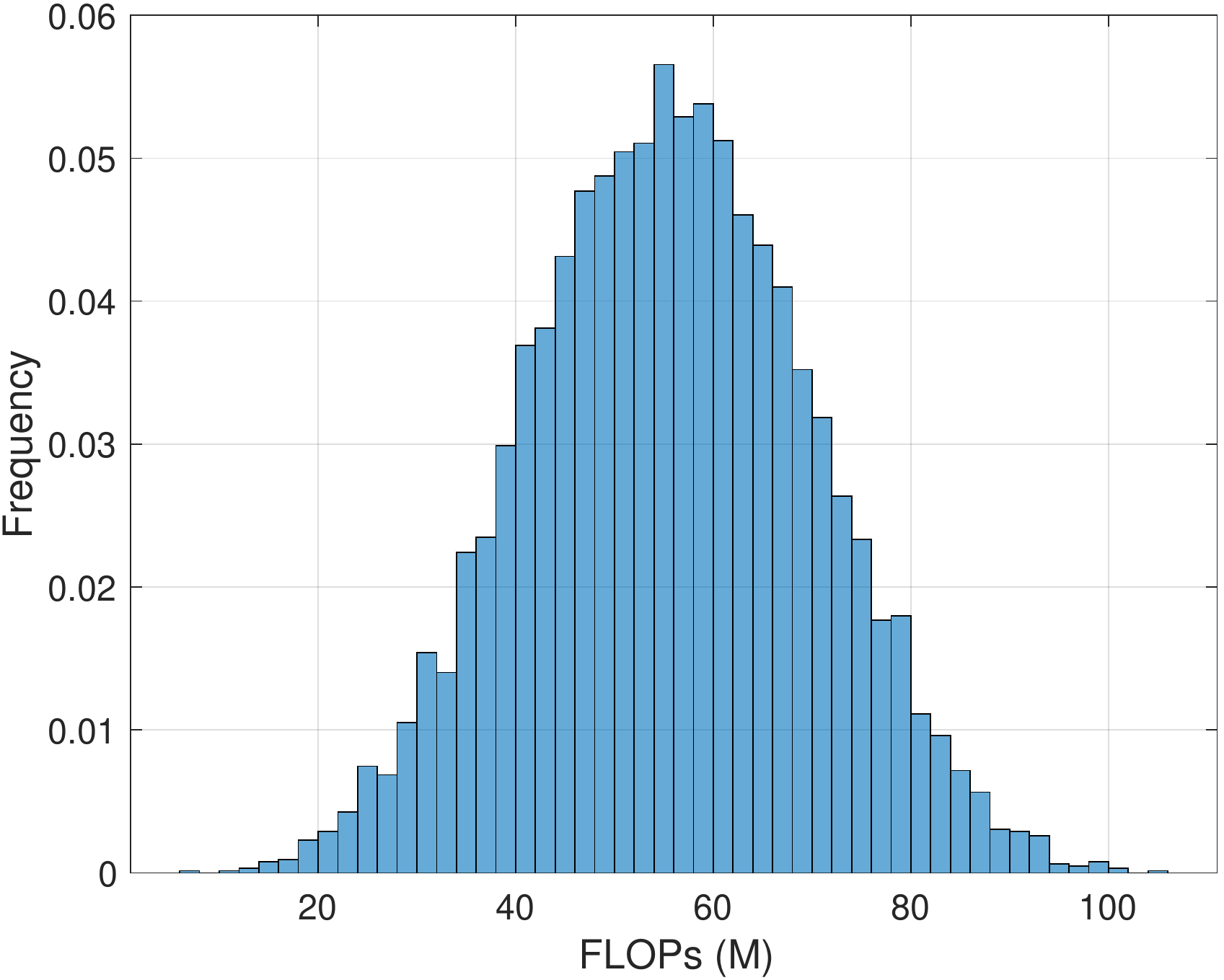}}
    \hspace{2mm}
    \subfigure[ACC]{\includegraphics[width=0.3\linewidth]{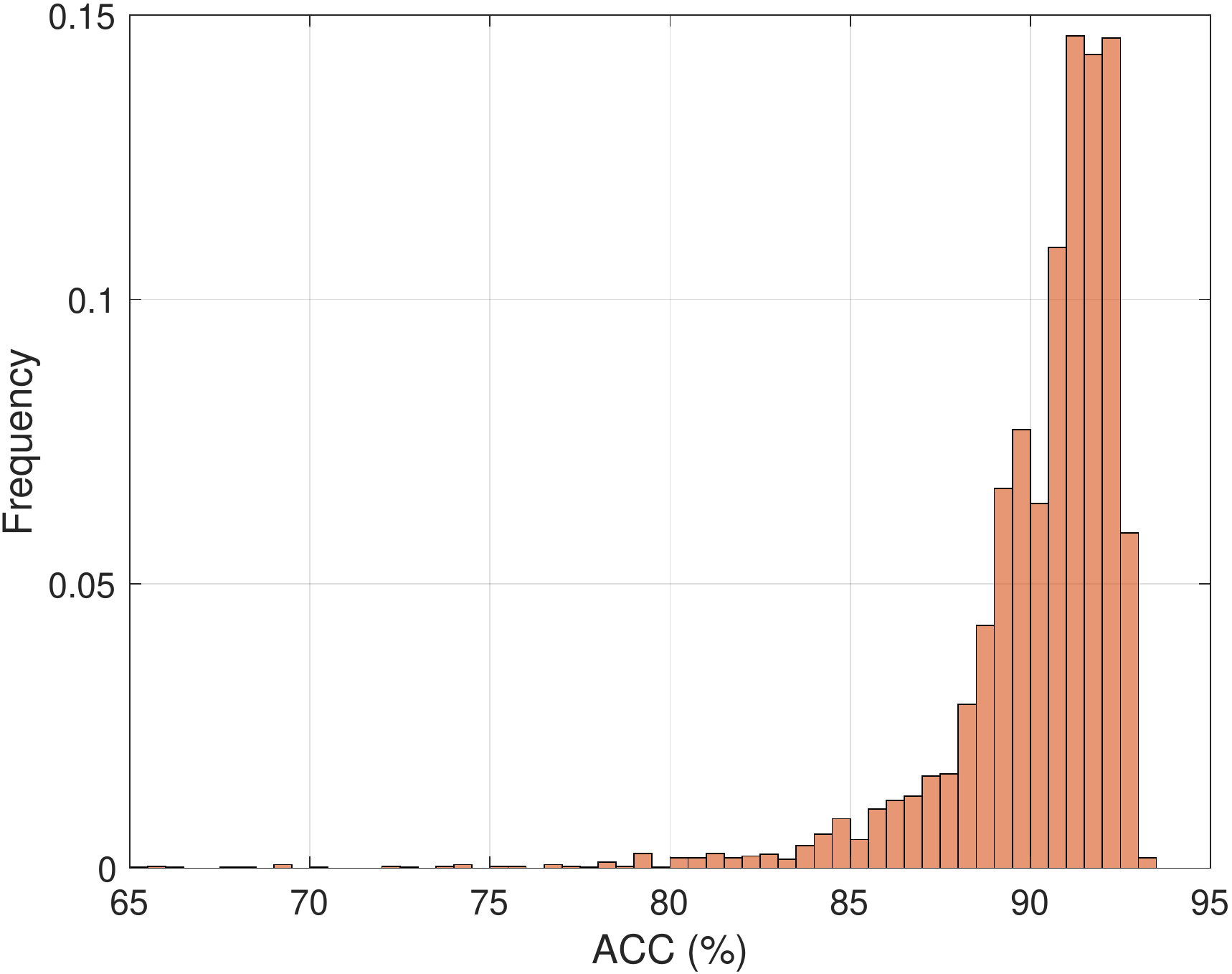}}
    \caption{Histograms of architectures in NAS-Bench-Macro \wrt (a) Params, (b) FLOPs, and (c) ACC.}
    \label{fig:nasbench_histograms}
\end{figure}

We analyze the distribution of parameters, FLOPs, and accuracies of all architectures in NAS-Bench-Macro, which are illustrated in Figure~\ref{fig:nasbench_histograms}.
From the results, we can infer that, although the FLOPs and parameters are distributed uniformly in the search space, but the accuracy performs differently, which might because when the capacity of model increases to a saturated level, the increment of capacity will not have remarkable performance increment; so the accuracies of a large number of architectures lie in a small range, this makes the NAS methods hard to rank those architectures accurately.

\subsection{Rank correlation of parameters and FLOPs with accuracies on NAS-Bench-Macro}
\label{sec:appendix:benchmark:rank}

We measure the rank correlation between parameters and accuracies, FLOPs and accuracies, respectively. The results summarized in Table~\ref{tab:nasbench_correlation_flops_params} show that the FLOPs has higher rank correlation coefficients than parameters. It indicates that the increment of FLOPs contributes more to accuracy than increasing parameters. 

\begin{table}[h]
	\centering
	\caption{Rank correlations of parameters and FLOPs with accuracies on NAS-Bench-Macro. }  
	\label{tab:nasbench_correlation_flops_params}
	\begin{tabular}{c||c|c}
		\hline
		type & Spearman rho (\%) & Kendall tau (\%) \\	
		\hline	
		params & 31.81 & 21.76 \\ 
		FLOPs & 66.09 & 55.60 \\
		\hline
	\end{tabular}
\end{table}

\subsection{Visualization of best architecture}
\begin{figure}[h]
    \centering
    \includegraphics[width=0.5\linewidth]{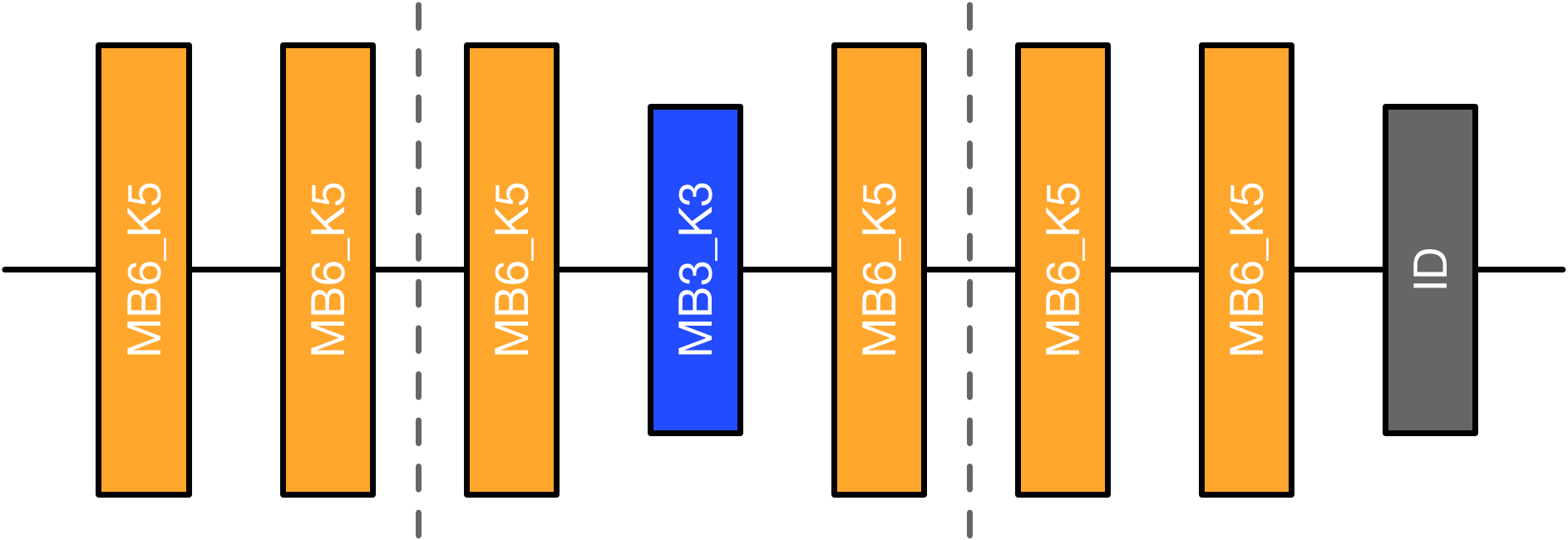}
    \caption{The best architecture on CIFAR-10 with 93.13\% test accuracy, 2.0M parameters and 85.2M FLOPs.}
    \label{fig:nasbench_vis_best}
\end{figure}

We visualize the best architecture of our NAS-Bench-Macro on CIFAR-10 dataset in Figure~\ref{fig:nasbench_vis_best}. The visualization shows that the architecture with the highest performance on CIFAR-10 tends to choose more large blocks (\textit{MB6\_K5}); however, an \textit{Identity} block on the last layer is used probably for easing the optimization pressure and decreasing the total receptive field.

\section{Details of Rank Correlation Coefficient}

    In Section \ref{sec:nasbench} and Section \ref{sec:appendix:benchmark:rank}, we rank the evaluation results of architectures on various validation sets with 1,000 and 50,000 samples. The obtained rankings are denoted as $\boldsymbol{r}$ and $\boldsymbol{s}$, respectively.
    To study the ranking consistency, we evaluate the correlation coefficient between them.
    Two metrics with different focuses are used, including the Kendall $\tau$ \cite{kendall1938new} and the Spearman $\rho$ \cite{pirie2004spearman} correlation coefficient.
    
    Firstly, the Kendall $\tau$ correlation coefficient focuses on the pairwise ranking performance. Considering a pair of index $i$ and $j$ such that $i < j$, if we have either both $\boldsymbol{r}_{i} > \boldsymbol{r}_{j}$ and $\boldsymbol{s}_{i} > \boldsymbol{s}_{j}$
    or both $\boldsymbol{r}_{i} < \boldsymbol{r}_{j}$ and $\boldsymbol{s}_{i} < \boldsymbol{s}_{j}$, the sort order of $(\boldsymbol{r}_{i}, \boldsymbol{r}_{j})$ and $(\boldsymbol{s}_{i}, \boldsymbol{s}_{j})$ agree, and pairs $(\boldsymbol{r}_{i}, \boldsymbol{s}_{i})$ and $(\boldsymbol{r}_{j}, \boldsymbol{s}_{j})$ are said to be \textit{concordant}.
    Otherwise, they disagree and the pairs are \textit{disconcordant}.
    With the concordant and disconcordant pairs, we can formally define the Kendall $\tau$ correlation coefficient by
    \begin{equation}
        \tau_K = \frac{N_{\text{concordant}} - N_{\text{disconcordant}}}{N_{\text{all}}},
        \label{eq:appendix:rank:kendall-tau-var}
    \end{equation}
    where $N_{\text{concordant}}$ and $N_{\text{disconcordant}}$ are the numbers of concordant and disconcordant pairs, respectively, and $\displaystyle N_{\text{all}} = \binom{n}{2} = \frac{n(n-1)}{2}$ is the total number of possible pairs out of $n$ overlapped elements.
    For efficient calculation, we can reformulate \eqref{eq:appendix:rank:kendall-tau-var} into an explicit expression form:
    \begin{equation}
        \tau_K = \frac{2}{n (n-1)} \sum_{i<j} \operatorname{sign} (\boldsymbol{r}_i - \boldsymbol{r}_j) \cdot \operatorname{sign} (\boldsymbol{s}_i - \boldsymbol{s}_j),
        \label{eq:appendix:rank:kendall-tau-vector}
    \end{equation}
    where $\operatorname{sign}(\cdot)$ is the sign function.
    
    Secondly, the Spearman $\rho$ correlation coefficient aims to evaluate to what degree a monotonic function fits the relationship between two random variables.
    If we consider $\boldsymbol{r}$ and $\boldsymbol{s}$ as two observation vectors of the random variables $r$ and $s$, respectively, the Spearman $\rho$ correlation coefficient can be calculated with the Pearson correlation coefficient:
    \begin{equation}
        \rho_S = \frac{\operatorname{cov}(r, s)}{\sigma_r \sigma_s},
        \label{eq:appendix:rank:spearman-rho-var}
    \end{equation}
    where $\operatorname{cov}(\cdot, \cdot)$ is the covariance of two variables, and $\sigma_r$ and $\sigma_s$ are the standard deviations of $r$ and $s$, respectively.
    In our experiment, the ranks are distinct integers. Therefore, Eq. \eqref{eq:appendix:rank:spearman-rho-var} can be reformulated as:
    \begin{equation}
        \rho_S = 1 - \frac{6 \sum_{i=1}^{n} (\boldsymbol{r}_i - \boldsymbol{s}_i)^2}{n (n^2 - 1)},
        \label{eq:appendix:rank:spearman-rho-vector}
    \end{equation}
    where $n=1000$ is the number of overlapped elements between $\boldsymbol{r}$ and $\boldsymbol{s}$.

\section{More Ablation Studies}

\subsection{Comparison between different training epochs on supernet}
In one-shot NAS, supernet as a performance evaluator highly correlates to the search performance. The more converged the supernet is, the more confident it will be on ranking subnets. However, increasing epochs of training also increase the computation cost; thus, we investigate the difference between different training epochs in this section. Concretely, we conduct experiments using our method with different training epochs on ImageNet, and measure their average validation accuracies and the corresponding train-from-scratch test accuracies of obtained architectures. The experiments are implemented with 330M FLOPs budget, and we search 20 subnets for each experiment.

\begin{figure}[h]
    \centering
    \includegraphics[width=0.46\linewidth]{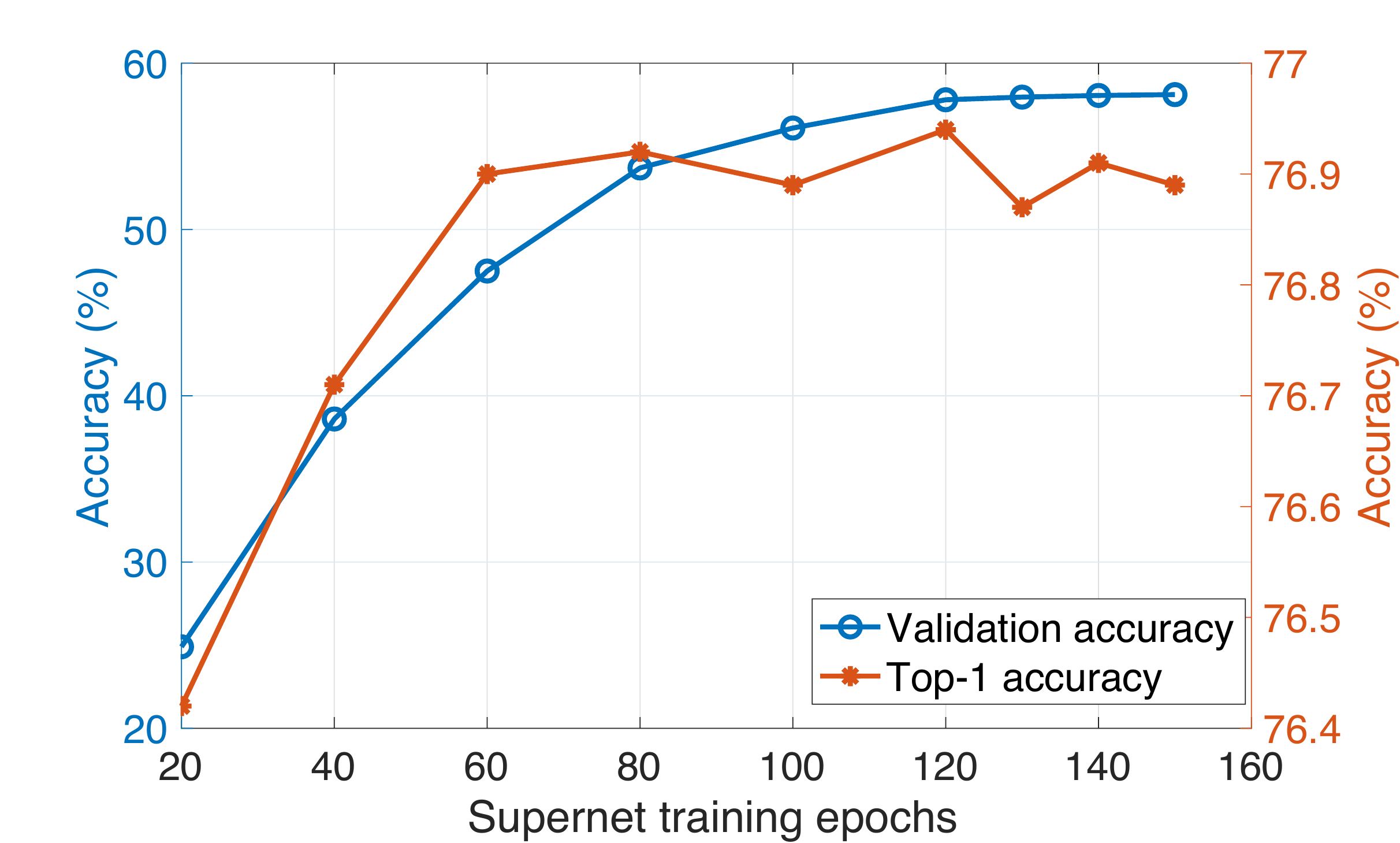}
    \caption{The Top-1 accuracy and average validation accuracy of searched architectures with different training epochs on supernet.}
    \label{fig:training_epochs}
\end{figure}

From the results illustrated in Figure \ref{fig:training_epochs}, 
we find the average validation accuracy of searched structures benefits from the increase of training epochs, and it almost achieves the peak after training epochs $N = 120$; moreover, we can find that when the training epochs $N=60$, the accuracy of obtained architecture is similar to the one with $N=120$.

\newpage
\subsection{Effect of the threshold constant $n_{thrd}$}
\begin{figure}[h]
    \centering
    \includegraphics[width=0.46\linewidth]{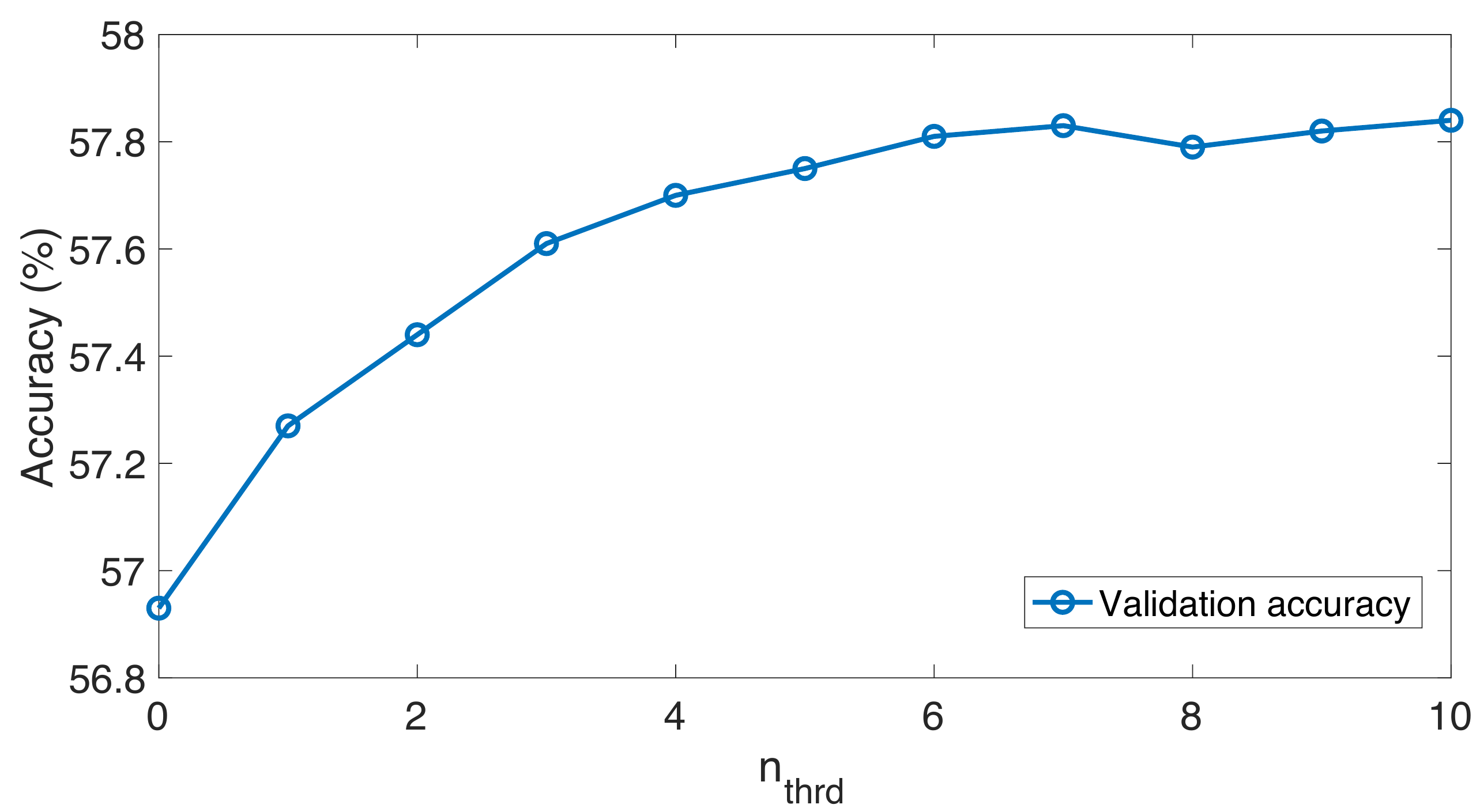}
    \caption{The validation accuracy of searched architectures with different threshold constant $n_{thrd}$.}
    \label{fig:nthrd}
    \vspace{-5mm}
\end{figure}

In MCT-NAS, the proposed hierarchical node selection aims to re-explore the less-visited nodes and evaluate the subnets more accurately with the defined threshold constant $n_{thrd}$. In detail, if the average number of visits of nodes in a layer is larger than $n_{thrd}$, we think it is promising to its reward, and thus the optimal node can be sampled with equation Eq. \eqref{eq:softmax-sampling} and Eq. \eqref{eq7}. To investigate the effect of the threshold constant $n_{thrd}$ to the validation accuracy of searched subnets, we conduct experiments to search with different $n_{thrd}$ on the same trained supernet of 330M FLOPs budget. In detail, we search 20 paths with different $n_{thrd}$ and record the average validation accuracy as Figure \ref{fig:nthrd}.

Figure \ref{fig:nthrd} shows that the larger $n_{thrd}$ leads to higher average validation accuracy of searched structures. Concretely, the accuracy of searched paths gradually increased as $n_{thrd}$ growing larger, which means that our proposed hierarchical node selection promotes to select for better paths. In addition, the performance tends to be stable when $n_{thrd}$ grows larger than 6, which implies $n_{thrd} = 6$ is sufficient to search for optimal paths.

\subsection{Trade-off between search number and $n_{thrd}$}

\begin{figure}[h]
    \centering
    \includegraphics[width=0.46\linewidth]{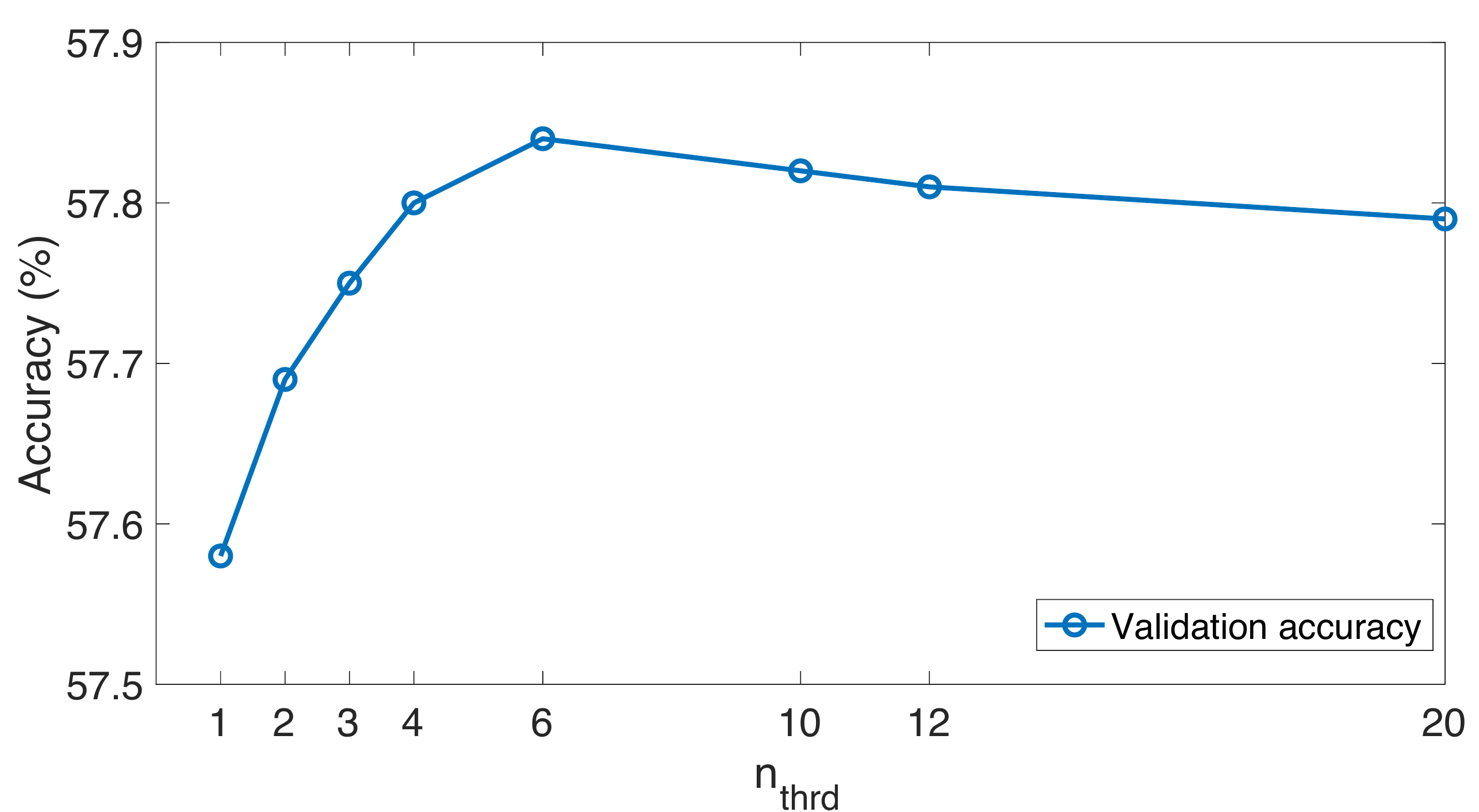}
    \caption{Average validation accuracy of searched architectures with different $n_{thrd}$ under the sane search cost.}
    \label{fig:trade_off}
    \vspace{-2mm}
\end{figure}

During the search, the increments of search number and $n_{thrd}$ both boost the search performance. However, under a certain upper limit of search cost, the search number should be in inversely proportional to the threshold constant $n_{thrd}$. Therefore, for a better trade-off between search number $S_n$ and $n_{thrd}$, we conduct experiments with different combinations of $S_n$ and $n_{thrd}$, which hold that 
\begin{equation} 
    S_n \times n_{thrd} = C
    \label{trade}
\end{equation} \normalsize
where $C$ is the target search cost coefficient, we implement the search with $C=120$ on the same trained supernet and report the average validation accuracy of searched architectures as in Figure \ref{fig:trade_off}.

In detail, we notice that when $n_{thrd}$ is small, the performance of the searched structure will increase with $n_{thrd}$ growing larger; this is because a larger $n_{thrd}$ promotes selecting nodes in MCT more accurately. Therefore, the performances of searched structures will be closer to the optimal one in the search space. However, when $n_{thrd}$ goes beyond 6, the accuracy of selected structures tend to decrease slightly; since with a default search budget, larger $n_{thrd}$ leads to a smaller search number, and thus it may hinder the search for the optimal structure.

\newpage
\subsection{Effect of MCT-NAS in supernet training}
\begin{figure}[h]
    \centering
    \includegraphics[width=0.48\linewidth]{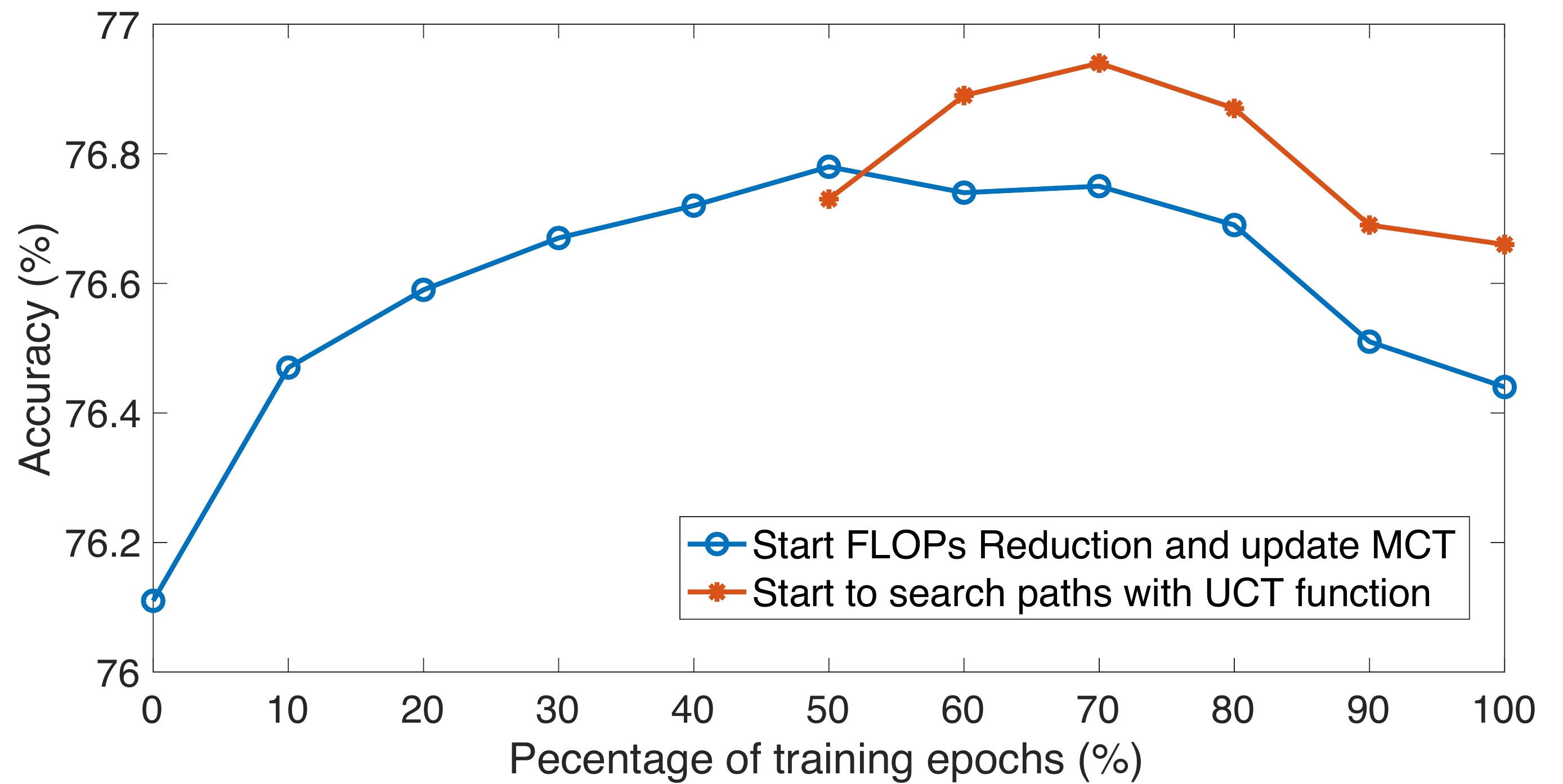}
    \caption{Effect of MCT-NAS in supernet training.}
    \label{fig:FLOPs_epochs}
\end{figure}

As illustrated in Algorithm \ref{supp_algorithm2}, we investigate two kinds of warm-ups in MCT-NAS, \ie, warm up of supernet and warm up of MCT. Besides, to investigate the ratio of iterations for these two terms (\ie, $W_S$ and $W_M$), we first conduct experiments with $W_S$  by setting $W_M$ to 0, \ie, we randomly uniform sample paths to update supernet and then select paths by FLOPs reduction and start to update MCT, as the \blue{blue line} shows in Figure \ref{fig:FLOPs_epochs}. In detail, we achieve the best performance when $W_S$ is set to 0.5. Afterward, we investigate the effect of $W_M$ with $W_S$ set to 0.5, as the \red{red line} shows in Figure \ref{fig:FLOPs_epochs}. With $W_S$ set to 0.2 (\ie, 70\% of training epochs), our searched structures achieve the best performance. As a result, with 50\% of overall training epochs, our supernet can effectively help to rank the performance of different structures; and then with 20\% training epochs, MCT is being well initialized and can be leveraged to effectively select paths in combination with UCT function (as defined in Eq.  \eqref{eq:softmax-sampling}, \eqref{eq:node-comm:uct-func-with-node-comm}). 
\vspace{-1mm}

\subsection{Effect of FLOPs reduction and updating MCT in training}
To examine the effect of FLOPs reduction and updating MCT, we implement supernet training under different settings of the proposed strategies in training. The detailed results can be referred to as Table \ref{MCT_NAS_analysis_supp}.

\begin{table*}[h]
	\caption{The performance gain of each part in MCT-NAS with 330M FLOPs on MobileNet search space.}
	\label{MCT_NAS_analysis_supp}
	\centering
	\scriptsize
	\begin{tabular}{|c|c|c|c|c|c|c|c|c||c|} \hline
		\multicolumn{5}{|c|}{Training} & \multicolumn{3}{c|}{Searching}&  \multicolumn{2}{c|}{Retraining} \\ \cline{1-10} 
		Uniform & FLOPs & Update MCT & UCT & Node & \multicolumn{1}{c|}{Evolutionary}& \multicolumn{1}{c}{MCTS} & Hierarchical & \multirow{2}*{Top-1} & \multirow{2}*{Top-5}\\ 
		Sampling &Reduction&in Training& Search & Communication &Search & Search  &  Updates &  &  \\ \hline 
		\checkmark& & & & & \checkmark & &  & 75.94\% & 92.89\% \\
		\checkmark& & & & &  & \checkmark & \checkmark & 76.44\% & 93.15\% \\
		\checkmark& \checkmark& & & & & \checkmark & \checkmark & 76.49\% & 93.19\% \\
		\checkmark& & \checkmark& & & & \checkmark & \checkmark & 76.54\% &     93.23\% \\
        \checkmark& \checkmark & \checkmark& & & &\checkmark & \checkmark & 76.65\% & 93.34\% \\
		\checkmark & \checkmark &\checkmark& & & & \checkmark & & 76.21\% & 93.11\%  \\
		\checkmark & \checkmark &\checkmark& \checkmark & & & \checkmark & & 76.35\% & 93.17\% \\ 
		\checkmark& \checkmark & \checkmark & \checkmark & \checkmark & &  \checkmark & & 76.62\% & 93.32\% \\
		\checkmark& \checkmark & \checkmark & \checkmark & \checkmark & &  \checkmark & \checkmark & 76.94\% & 93.37\% \\ \hline
	\end{tabular} 
	\vspace{-2mm}
\end{table*}
In detail, we can see that with MCT sampler, FLOPs reduction (updating MCT in training) can lead to a 0.05\%(0.1\%) improvement on Top-1 accuracy ($3$-th and $4$-th row in Table \ref{MCT_NAS_analysis_supp}). Moreover, when use both strategies of FLOPs reduction and updating MCT in training ($5$-th row in Table \ref{MCT_NAS_analysis_supp}), the performance of searched results can be boosted by 0.21\%; Since with FLOPs Reduction, each sampled subnet is within the FLOPs budget, and thus can help to efficiently update MCT.

\newpage
\section{The Calculation of Search Cost in MCT-NAS}
In this paper, we propose a hierarchical node selection method, which involves additional computation cost. For an accurate and fair comparison with other NAS methods, we calculate the total computed image number for the replacement of the search number.
Many methods leverage the validation dataset \cite{guo2020single,you2020greedynas} (split from training) with 50k pictures to evaluate each path. However, in MCT-NAS, since we evaluate each path with a single batch of pictures, the number of pictures required for selecting a proper path is formulated as follows
\begin{equation}
    \mathcal{N}_{path} = bs \times n_{thrd} \times \sum_{i=1}^{L} |\mathcal{O}_i|
    \label{N_path}
\end{equation}
where $bs$ represents the batch size in search, and $|\mathcal{O}_i|$ indicates the number of candidate operations in layer $i$. Therefore, with $bs=128$, $n_{thrd} =6$, $L=21$, and $|\mathcal{O}_i|=13$, each path in MCT-NAS amounts to $5\times$ search cost compared to other search methods.

\section{Visualization of Searched Architectures}
\begin{figure}
    \centering
    \subfigure[MCT-NAS-A]{\includegraphics[width=0.20\linewidth]{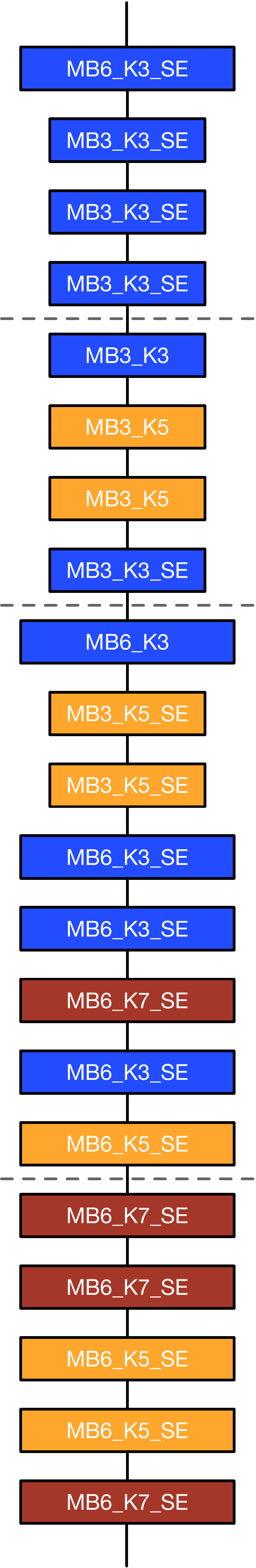}}
    \hspace{5mm}
    \subfigure[MCT-NAS-B]{\includegraphics[width=0.20\linewidth]{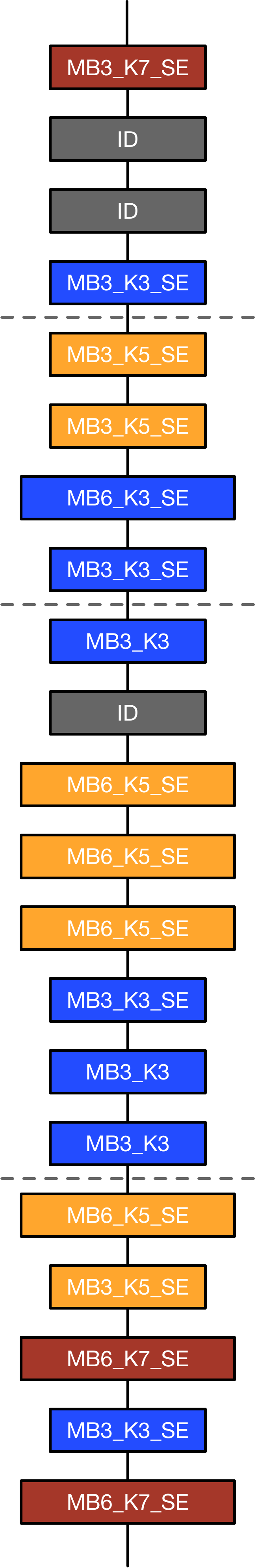}}
    \hspace{5mm}
    \subfigure[MCT-NAS-C]{\includegraphics[width=0.20\linewidth]{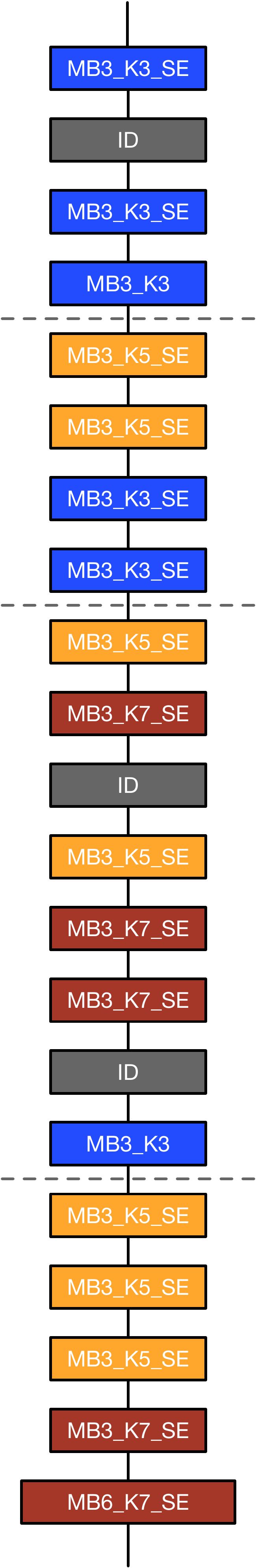}}
    \caption{Visualization of architectures obtained by MCT-NAS in Table \ref{tbl:sota}.}
    \label{fig:sota_vis}
\end{figure}

We visualize the searched architectures by our MCTS-NAS method as Figure \ref{fig:sota_vis}. In detail, we find that the searched optimal structures have the following three characteristics:

\begin{enumerate}
	
	\item In the kernel level, the optimal structure generally tends to use more 5$\times$5 or 7$\times$7 convolutions at the layers close to the last layer, and the last layer is always with 7$\times$7 kernel size.
	
	\item In the width level, the last few operations of the searched structure are more likely to use a large expansion ratio, which is more evident with a larger FLOPs budget (\eg, our 440M FLOPs structure uses an expansion ratio of 6 for the last 10 consecutive layers).
	
	\item When the network budget is insufficient (\ie 330M and 280M), the searched structure generally tends to use more ID operations at the layers close to the first layer to reduce FLOP consumption through fast downsampling.
	
\end{enumerate}

\end{document}